\documentclass[preprint]{elsarticle}

\usepackage{hyperref}
\usepackage{amsmath,amsfonts,bm}
\usepackage{booktabs}
  \usepackage[table,xcdraw]{xcolor}
\usepackage{xcolor}
\usepackage{subcaption}
\usepackage{enumitem}
\usepackage{wrapfig}
\usepackage{caption}

\usepackage[normalem]{ulem}
\useunder{\uline}{\ul}{}

\usepackage{ifthen}
\newboolean{anon}  
\setboolean{anon}{false}

\begin{document}
% Color Commands
%\newcommand{green}[1] {{\color{green} {#1}}}

%--------------------------------------------------------------------
\begin{frontmatter}
\title{Temporal Fusion Transformers \\for Interpretable Multi-horizon Time Series Forecasting}

\ifthenelse{\boolean{anon}}{\author{Anonymous Authors}}{
\author[oxf]{Bryan Lim\corref{mycorrespondingauthor}\fnref{myfootnote}}
\fntext[myfootnote]{Completed as part of internship with Google Cloud AI Research.}
\cortext[mycorrespondingauthor]{Corresponding authors}
\ead{blim@robots.ox.ac.uk}
\author[goog]{Sercan \"{O}. Ar{\i}k}
\ead{soarik@google.com}
\author[goog]{Nicolas Loeff}
\ead{nloeff@google.com}
\author[goog]{Tomas Pfister}
\ead{tpfister@google.com}
\address[oxf]{University of Oxford, UK}
\address[goog]{Google Cloud AI, USA}
}


\begin{abstract}
Multi-horizon forecasting often contains a complex mix of inputs -- including static (i.e. time-invariant) covariates, known future inputs, and other exogenous time series that are only observed in the past -- without any prior information on how they interact with the target. 
Several deep learning methods have been proposed, but they are typically `black-box' models which do not shed light on how they use the full range of inputs present in practical scenarios. 
In this paper, we introduce the Temporal Fusion Transformer (TFT) -- a novel attention-based architecture which combines high-performance multi-horizon forecasting with interpretable insights into temporal dynamics. 
To learn temporal relationships at different scales, TFT uses recurrent layers for local processing and interpretable self-attention layers for long-term dependencies.
TFT utilizes specialized components to select relevant features and a series of gating layers to suppress unnecessary components, enabling high performance in a wide range of scenarios. 
On a variety of real-world datasets, we demonstrate significant performance improvements over existing benchmarks, and showcase three practical interpretability use cases of TFT. 
\end{abstract}

\begin{keyword}
Deep learning, Interpretability, Time series, Multi-horizon forecasting, Attention mechanisms, Explainable AI.
\end{keyword}

\end{frontmatter}

%\linenumbers
\section{Introduction}

%MHF definition & why it matters
Multi-horizon forecasting, i.e. the prediction of variables-of-interest at multiple future time steps, is a crucial problem within time series machine learning.  
In contrast to one-step-ahead predictions, multi-horizon forecasts provide users with access to estimates across the entire path, allowing them to optimize their actions at multiple steps in future (e.g. retailers optimizing the inventory for the entire upcoming season, or clinicians optimizing a treatment plan for a patient). 
Multi-horizon forecasting has many impactful real-world applications in retail \cite{RetailDemandExample, FashionExample}, healthcare \cite{RMSN, EpidemiologyExample} and economics \cite{CAPISTRAN2010666}) -- performance improvements to existing methods in such applications are highly valuable.

\begin{figure}[!htbp]
  \centering
  \includegraphics[width=0.6\linewidth]{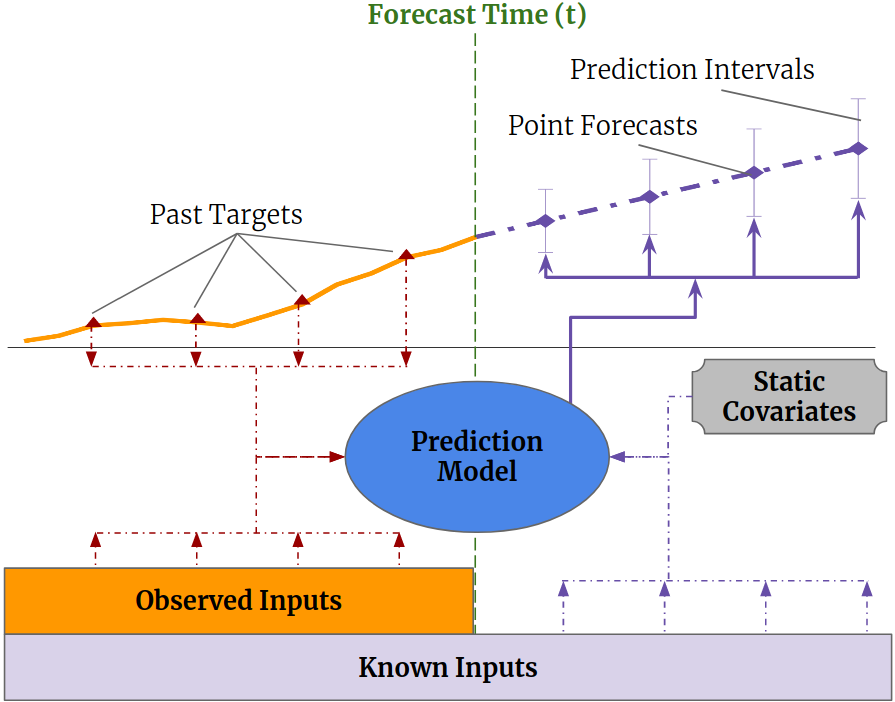}
  \caption{Illustration of multi-horizon forecasting with static covariates, past-observed and apriori-known future time-dependent inputs.}
  \label{fig:multihorizonforecasting}
\end{figure}
Practical multi-horizon forecasting applications commonly have access to a variety of data sources, as shown in Fig. \ref{fig:multihorizonforecasting}, including known information about the future (e.g. upcoming holiday dates), other exogenous time series (e.g. historical customer foot traffic), and static metadata (e.g. location of the store) -- without any prior knowledge on how they interact.
This heterogeneity of data sources together with little information about their interactions makes multi-horizon time series forecasting particularly challenging.  

Deep neural networks (DNNs) have increasingly been used in multi-horizon forecasting, demonstrating strong performance improvements over traditional time series models \cite{DSSM,AttentionDiseaseProgression, M4Competition}. 
While many architectures have focused on variants of recurrent neural network (RNN) architectures \cite{DeepAR, DSSM, MQRNN}, recent improvements have also used attention-based methods to enhance the selection of relevant time steps in the past \cite{JDLstm} -- including Transformer-based models \cite{LiTransformer}. 
However, these often fail to consider the different types of inputs commonly present in multi-horizon forecasting, and either assume that all exogenous inputs are known into the future \cite{DeepAR, DSSM, LiTransformer} -- a common problem with autoregressive models -- or neglect important static covariates \cite{MQRNN} -- which are simply concatenated with other time-dependent features at each step.
Many recent improvements in time series models have resulted from the alignment of architectures with unique data characteristics \cite{ClockworkRNN, PhasedLSTM}. 
We argue and demonstrate that similar performance gains can also be reaped by designing networks with suitable inductive biases for multi-horizon forecasting.

% Challenge 2
In addition to not considering the heterogeneity of common multi-horizon forecasting inputs, most current architectures are `black-box' models where forecasts are controlled by complex nonlinear interactions between many parameters.
This makes it difficult to explain how models arrive at their predictions, and in turn makes it challenging for users to trust a model's outputs and model builders to debug it. 
Unfortunately, commonly-used explainability methods for DNNs are not well-suited for applying to time series.
In their conventional form, post-hoc methods (e.g. LIME \cite{LIME} and SHAP \cite{SHAP}) do not consider the time ordering of input features.
For example, for LIME, surrogate models are independently constructed for each data-point, and for SHAP, features are considered independently for neighboring time steps. 
Such post-hoc approaches would lead to poor explanation quality as dependencies between time steps are typically significant in time series. 
On the other hand, some attention-based architectures are proposed with inherent interpretability for sequential data, primarily language or speech -- such as the Transformer architecture \cite{Transformer}. The fundamental caveat to apply them is that multi-horizon forecasting includes many different types of input features, as opposed to language or speech.
In their conventional form, these architectures can provide insights into relevant time steps for multi-horizon forecasting, but they cannot distinguish the importance of different features at a given timestep.
Overall, in addition to the need for new methods to tackle the heterogeneity of data in multi-horizon forecasting for high performance, new methods are also needed to render these forecasts interpretable, given the needs of the use cases.

%This does not enable understanding how static (i.e. time-invariant) covariates play a key role in predictions, for example in healthcare where genetic information can often determine the expression of a disease \cite{GeneExample}. 

% Solution
In this paper we propose the Temporal Fusion Transformer (TFT) -- an attention-based DNN architecture for multi-horizon forecasting that achieves high performance while enabling new forms of interpretability.
To obtain significant performance improvements over state-of-the-art benchmarks, we introduce multiple novel ideas to align the architecture with the full range of potential inputs and temporal relationships common to multi-horizon forecasting -- specifically incorporating 
(1) static covariate encoders which encode context vectors for use in other parts of the network, 
(2) gating mechanisms throughout and sample-dependent variable selection to minimize the contributions of irrelevant inputs, 
(3) a sequence-to-sequence layer to locally process known and observed inputs, and 
(4) a temporal self-attention decoder to learn any long-term dependencies present within the dataset. 
The use of these specialized components also facilitates interpretability; in particular, we show that TFT enables three valuable interpretability use cases: helping users identify 
(i) globally-important variables for the prediction problem, 
(ii) persistent temporal patterns, and 
(iii) significant events. 
On a variety of real-world datasets, we demonstrate how TFT can be practically applied, as well as the insights and benefits it provides.

\section{Related Work}
\label{sec:related_works}
 
\textbf{DNNs for Multi-horizon Forecasting:} Similarly to traditional multi-horizon forecasting methods \cite{MultistepMethods, DirectVsIterated}, recent deep learning methods can be categorized into iterated approaches using autoregressive models \cite{DeepAR, DSSM, LiTransformer} or direct methods based on sequence-to-sequence models \cite{MQRNN, JDLstm}.   

Iterated approaches utilize one-step-ahead prediction models, with multi-step predictions obtained by recursively feeding predictions into future inputs. 
Approaches with Long Short-term Memory (LSTM) \cite{LSTM} networks have been considered, such as Deep AR \cite{DeepAR} which uses stacked LSTM layers to generate parameters of one-step-ahead Gaussian predictive distributions. 
Deep State-Space Models (DSSM) \cite{DSSM} adopt a similar approach, utilizing LSTMs to generate parameters of a predefined linear state-space model with predictive distributions produced via Kalman filtering -- with extensions for multivariate time series data in \cite{DeepFactorsForForecasting}. 
More recently, Transformer-based architectures have been explored in \cite{LiTransformer}, which proposes the use of convolutional layers for local processing and a sparse attention mechanism to increase the size of the receptive field during forecasting. 
Despite their simplicity, iterative methods rely on the assumption that the values of all variables excluding the target are known at forecast time -- such that only the target needs to be recursively fed into future inputs. 
However, in many practical scenarios, numerous useful time-varying inputs exist, with many unknown in advance.
Their straightforward use is hence limited for iterative approaches. 
TFT, on the other hand, explicitly accounts for the diversity of inputs -- naturally handling static covariates and (past-observed and future-known) time-varying inputs.

In contrast, direct methods are trained to explicitly generate forecasts for multiple predefined horizons at each time step. 
Their architectures typically rely on sequence-to-sequence models, e.g. LSTM encoders to summarize past inputs, and a variety of methods to generate future predictions. 
The Multi-horizon Quantile Recurrent Forecaster (MQRNN) \cite{MQRNN} uses LSTM or convolutional encoders to generate context vectors which are fed into multi-layer perceptrons (MLPs) for each horizon. 
In \cite{JDLstm} a multi-modal attention mechanism is used with LSTM encoders to construct context vectors for a bi-directional LSTM decoder.
Despite performing better than LSTM-based iterative methods, interpretability remains challenging for such standard direct methods. 
In contrast, we show that by interpreting attention patterns, TFT can provide insightful explanations about temporal dynamics, and do so while maintaining state-of-the-art performance on a variety of datasets.

\textbf{Time Series Interpretability with Attention:} Attention mechanisms are used in translation \cite{Transformer}, image classification \cite{ImageAttentionEx} or tabular learning \cite{arik2019tabnet} to identify salient portions of input for each instance using the magnitude of attention weights. 
Recently, they have been adapted for time series with interpretability motivations \cite{AttentionDiseaseProgression,LiTransformer, RETAIN}, using LSTM-based \cite{AttendAndDiagnose} and transformer-based \cite{LiTransformer} architectures.
However, this was done without considering the importance of static covariates (as the above methods blend variables at each input). 
TFT alleviates this by using separate encoder-decoder attention for static features at each time step on top of the self-attention to determine the contribution time-varying inputs. 

\textbf{Instance-wise Variable Importance with DNNs: }
Instance (i.e. sample)-wise variable importance can be obtained with post-hoc explanation methods \cite{LIME,SHAP, yoon2019rllim} and inherently intepretable models \cite{LSTMVariableImportance,RETAIN}. 
Post-hoc explanation methods, e.g. LIME \cite{LIME}, SHAP \cite{SHAP} and RL-LIM \cite{yoon2019rllim}, are applied on pre-trained black-box models and often based on distilling into a surrogate interpretable model, or decomposing into feature attributions. 
They are not designed to take into account the time ordering of inputs, limiting their use for complex time series data. 
Inherently-interpretable modeling approaches build components for feature selection directly into the architecture. 
For time series forecasting specifically, they are based on explicitly quantifying time-dependent variable contributions. For example, Interpretable Multi-Variable LSTMs \cite{LSTMVariableImportance} partitions the hidden state such that each variable contributes uniquely to its own memory segment, and weights memory segments to determine variable contributions. 
Methods combining temporal importance and variable selection have also been considered in \cite{RETAIN}, which computes a single contribution coefficient based on attention weights from each. 
However, in addition to the shortcoming of modelling only one-step-ahead forecasts, existing methods also focus on \textit{instance-specific} (i.e. sample-specific) interpretations of attention weights -- without providing insights into global temporal dynamics. 
In contrast, the use cases in Sec. \ref{sec:interp} demonstrate that TFT is able to analyze global temporal relationships and allows users to interpret global behaviors of the model on the whole dataset -- specifically in the identification of any persistent patterns (e.g. seasonality or lag effects) and regimes present.
\section{Multi-horizon Forecasting}
\label{sec:problem_def}

%The general problem of multi-horizon forecasting is depicted in Fig. \ref{fig:multihorizonforecasting}. 
Let there be $I$ unique entities in a given time series dataset -- such as different stores in retail or patients in healthcare. Each entity $i$ is associated with a set of static covariates $\bm{s}_i \in \mathbb{R}^{m_s}$, as well as inputs $\bm{\chi}_{i,t} \in \mathbb{R}^{m_\chi}$ and scalar targets $y_{i,t} \in \mathbb{R}$ at each time-step $t \in [0, T_i]$. 
Time-dependent input features are subdivided into two categories $\bm{\chi}_{i,t} = \left[\bm{z}_{i,t}^T, \bm{x}_{i,t}^T\right]^T$  -- observed inputs $\bm{z}_{i,t} \in \mathbb{R}^{ (m_z)}$ which can only be measured at each step and are unknown beforehand, and known inputs $\bm{x}_{i,t} \in \mathbb{R}^{m_x}$ which can be predetermined (e.g. the day-of-week at time $t$). 

In many scenarios, the provision for prediction intervals can be useful for optimizing decisions and risk management by yielding an indication of likely best and worst-case values that the target can take. 
As such, we adopt quantile regression to our multi-horizon forecasting setting (e.g. outputting the $10^{th}$, $50^{th}$ and $90^{th}$ percentiles at each time step). 
Each quantile forecast takes the form:
\begin{equation}
\hat{y}_{i}(q,t, \tau) = f_q\left(\tau, y_{i,t-k:t}, \bm{z}_{i,t-k:t}, \bm{x}_{i,t-k:t+\tau}, \bm{s}_i\right), 
\end{equation}
where $\hat{y}_{i,t+\tau}(q,t ,\tau)$ is the predicted $q^{th}$ sample quantile of the $\tau$-step-ahead forecast at time $t$, and $f_q(.)$ is a prediction model. In line with other direct methods, we simultaneously output forecasts for $\tau_{max}$ time steps -- i.e. $\tau \in \{1, \dots, \tau_{max}\}$. 
We incorporate all past information within a finite look-back window $k$, using target and known inputs only up till and including the forecast start time $t$ (i.e. $y_{i,t-k:t} = \left\{y_{i,t-k}, \dots, y_{i,t} \right\}$) and known inputs across the entire range (i.e. $\bm{x}_{i,t-k:t+\tau} = \big\{\bm{x}_{i,t-k}, \dots,$ $\bm{x}_{i,t}, \dots, \bm{x}_{i,t+\tau} \big\}$). \footnote{For notation simplicity, we omit the subscript $i$ unless explicitly required.}

\section{Model Architecture}

\begin{figure}[htb]
  \centering
  \includegraphics[width=1.0\linewidth]{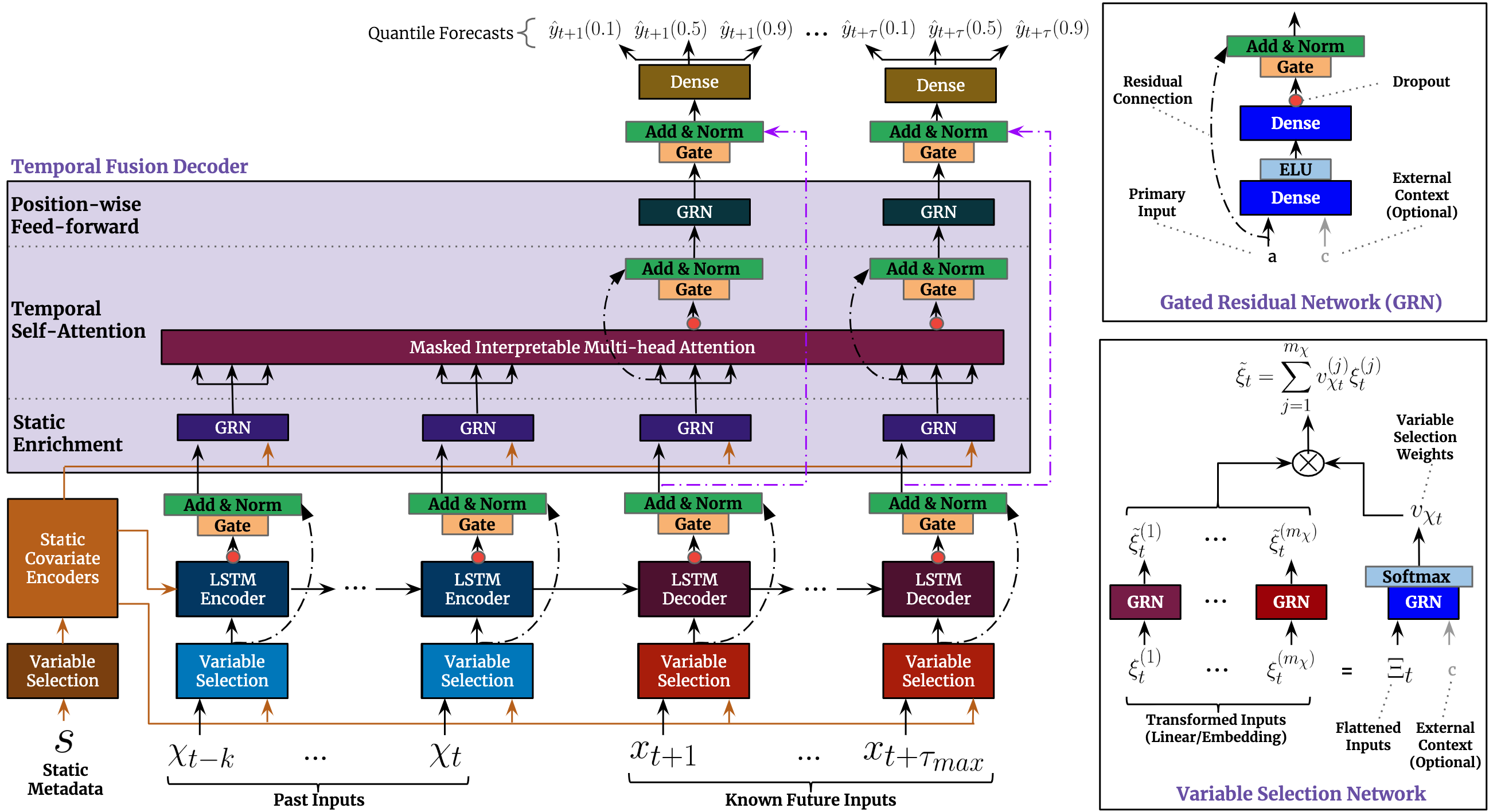}
  \caption{TFT architecture. TFT inputs static metadata, time-varying past inputs and time-varying a priori known future inputs. 
  Variable Selection is used for judicious selection of the most salient features based on the input. 
  Gated Residual Network blocks enable efficient information flow with skip connections and gating layers.
  Time-dependent processing is based on LSTMs for local processing, and multi-head attention for integrating information from any time step.}
  \label{fig:tft_architecture}
\end{figure}

We design TFT to use canonical components to efficiently build feature representations for each input type (i.e. static, known, observed inputs) for high forecasting performance on a wide range of problems. The major constituents of TFT are:
\begin{enumerate}[noitemsep, nolistsep, leftmargin=*]
\item \textbf{Gating mechanisms} to skip over any unused components of the architecture, providing adaptive depth and network complexity to accommodate a wide range of datasets and scenarios. 
%Gated Linear Units extensively are utilized throughout our architecture, and Gated Residual Network is proposed as a main building block.
\item \textbf{Variable selection networks} to select relevant input variables at each time step.
\item \textbf{Static covariate encoders} to integrate static features into the network, through encoding of context vectors to condition temporal dynamics.
\item \textbf{Temporal processing} to learn both long- and short-term temporal relationships from both observed and known time-varying inputs. A sequence-to-sequence layer is employed for local processing, whereas long-term dependencies are captured using a novel interpretable multi-head attention block.
\item \textbf{Prediction intervals} via quantile forecasts to determine the range of likely target values at each prediction horizon.
\end{enumerate}

Fig. \ref{fig:tft_architecture} shows the high level architecture of Temporal Fusion Transformer (TFT), with individual components described in detail in the subsequent sections. 

\subsection{Gating Mechanisms}
The precise relationship between exogenous inputs and targets is often unknown in advance, making it difficult to anticipate which variables are relevant. Moreover, it is difficult to determine the extent of required non-linear processing, and there may be instances where simpler models can be beneficial -- e.g. when datasets are small or noisy.
With the motivation of giving the model the flexibility to apply non-linear processing only where needed, we propose Gated Residual Network (GRN) as shown in in Fig. \ref{fig:tft_architecture} as a building block of TFT.
The GRN takes in a primary input $\bm{a}$ and an optional context vector $\bm{c}$ and yields:
\begin{align}
\text{GRN}_\omega\left(\bm{a}, \bm{c} \right) &=\text{LayerNorm}\left(\bm{a} + \text{GLU}_\omega(\bm{\eta}_1) \right), \\
    \bm{\eta}_1 &= \bm{W}_{1, \omega}~\bm{\eta}_2 + \bm{b}_{1, _\omega},  \label{eqn:grn_intermediate}\\
    \bm{\eta}_2 &= \text{ELU}\left( \bm{W}_{2, \omega}~\bm{a} + \bm{W}_{3, \omega}~\bm{c}  + \bm{b}_{2, _\omega} \right), & \label{eqn:grn_base}
\end{align}
where $\text{ELU}$ is the Exponential Linear Unit activation function \cite{elu}, $\bm{\eta}_1 \in \mathbb{R}^{d_{model}}, \bm{\eta}_2  \in \mathbb{R}^{d_{model}}$ are intermediate layers, $\text{LayerNorm}$ is standard layer normalization of \cite{LayerNorm}, and $\omega$ is an index to denote weight sharing. When $\bm{W}_{2, \omega}~\bm{a} + \bm{W}_{3, \omega}~\bm{c}  + \bm{b}_{2, _\omega} >> 0$, the ELU activation would act as an identity function and when $\bm{W}_{2, \omega}~\bm{a} + \bm{W}_{3, \omega}~\bm{c}  + \bm{b}_{2, _\omega} << 0$, the ELU activation would generate a constant output, resulting in linear layer behavior.
%On top of the variable selection described in Sec. \ref{sec:variable_selection},
We use component gating layers based on Gated Linear Units (GLUs) \citep{GLU} to provide the flexibility to suppress any parts of the architecture that are not required for a given dataset. Letting $\bm{\gamma} \in \mathbb{R}^{d_{model}}$ be the input, the GLU then takes the form:
\begin{align}
 \text{GLU}_\omega(\bm{\gamma}) & =  \sigma(\bm{W}_{4, \omega}~\bm{\gamma} + \bm{b}_{4, \omega}) \odot (\bm{W}_{5, \omega}~\bm{\gamma} + \bm{b}_{5, \omega} ),
\label{eqn:component_gate}
\end{align}
where $\sigma(.)$ is the sigmoid activation function, 
%which controls a component's transformed output by gating,
$\bm{W}_{(.)} \in \mathbb{R}^{d_{model}\times d_{model}}$, $\bm{b}_{(.)} \in \mathbb{R}^{d_{model}}$ are the weights and biases, $\odot$ is the element-wise Hadamard product, and $d_{model}$ is the hidden state size (common across TFT).
GLU allows TFT to control the extent to which the GRN contributes to the original input $\bm{a}$ -- potentially skipping over the layer entirely if necessary as the GLU outputs could be all close to 0 in order to surpress the nonlinear contribution. 
For instances without a context vector, the GRN simply treats the contex input as zero -- i.e. $\bm{c} = 0$ in Eq. \eqref{eqn:grn_base}. During training, dropout is applied before the gating layer and layer normalization -- i.e. to $\bm{\eta}_1$ in Eq. \eqref{eqn:grn_intermediate}.

\subsection{Variable Selection Networks}
\label{sec:variable_selection}
While multiple variables may be available, their relevance and specific contribution to the output are typically unknown. 
TFT is designed to provide instance-wise variable selection through the use of variable selection networks applied to both static covariates and time-dependent covariates. 
Beyond providing insights into which variables are most significant for the prediction problem, variable selection also allows TFT to remove any unnecessary noisy inputs which could negatively impact performance. Most real-world time series datasets contain features with less predictive content, thus variable selection can greatly help model performance via utilization of learning capacity only on the most salient ones.

We use entity embeddings \cite{EntityEmbeddings} for categorical variables as feature representations, and linear transformations for continuous variables -- transforming each input variable into a $(d_{model})$-dimensional vector which matches the dimensions in subsequent layers for skip connections. 
All static, past and future inputs make use of separate variable selection networks (as denoted by different colors in Fig. \ref{fig:tft_architecture}). 
Without loss of generality, we present the variable selection network for past inputs -- noting that those for other inputs take the same form. 

Let $\bm{\xi}_{t}^{(j)}  \in \mathbb{R}^{d_{model}}$ denote the transformed input of the $j$-th variable at time $t$, with $\bm{\Xi}_t = \left[ \bm{\xi}_{t}^{(1)^T}, \dots, \bm{\xi}_{t}^{(m_\chi)^T}  \right]^T$ being the flattened vector of all past inputs at time $t$. Variable selection weights are generated by feeding both $\bm{\Xi}_t$ and an external context vector $\bm{c}_{s}$ through a GRN, followed by a Softmax layer: 
\begin{align}
\bm{v}_{\chi_t} = \text{Softmax}\left(\text{GRN}_{v_{\chi}}(\bm{\Xi}_t, \bm{c}_{s})\right),& 
\label{eq:var_selection}
\end{align}
where $\bm{v}_{\chi_t} \in \mathbb{R}^{m_\chi}$ is a vector of variable selection weights, and $\bm{c}_{s}$ is obtained from a static covariate encoder (see Sec. \ref{sec:static_cov_encoder}). For static variables, we note that the context vector $\bm{c}_{s}$ is omitted -- given that it already has access to static information.

At each time step, an additional layer of non-linear processing is employed by feeding each $\bm{\xi}_{t}^{(j)}$ through its own GRN:
\begin{align}
\tilde{\bm{\xi}}_t^{(j)} = \text{GRN}_{\tilde{\xi}(j)}\left(\bm{\xi}_t^{(j)}\right), & \label{sec:varselect_grn}
\end{align}
where $\tilde{\bm{\xi}}_t^{(j)}$ is the processed feature vector for variable $j$. We note that each variable has its own $\text{GRN}_{\xi(j)}$, with \emph{weights shared across all time steps $t$}. Processed features are then weighted by their variable selection weights and combined:
\begin{align}
\tilde{\bm{\xi}}_t  = \sum\nolimits_{j=1}^{m_\chi} v_{\chi_t}^{(j)}  \tilde{\bm{\xi}}_t^{(j)}, &
\label{eq:var_selection_sum}
\end{align}
where $v_{\chi_t}^{(j)}$ is the j-th element of vector  $\bm{v}_{\chi_t} $. 
 
\subsection{Static Covariate Encoders}
\label{sec:static_cov_encoder}
In contrast with other time series forecasting architectures, the TFT is carefully designed to integrate information from static metadata, using separate GRN encoders to produce four different context vectors, $\bm{c}_s$, $\bm{c}_e$, $\bm{c}_c$, and $\bm{c}_h$. These contect vectors are wired into various locations in the temporal fusion decoder (Sec. \ref{sec:temporal_fusion}) where static variables play an important role in processing. Specifically, this includes contexts for 
(1) temporal variable selection ($\bm{c}_{s}$), 
(2) local processing of temporal features ($\bm{c}_{c}, \bm{c}_{h}$), and 
(3) enriching of temporal features with static information ($\bm{c}_{e})$. 
As an example, taking $\bm{\zeta}$ to be the output of the static variable selection network, contexts for temporal variable selection would be encoded according to $\bm{c}_{s} = GRN_{c_s}(\bm{\zeta})$.

\subsection{Interpretable Multi-Head Attention}
\label{sec:interp_multi_head}
The TFT employs a self-attention mechanism to learn long-term relationships across different time steps, which we modify from multi-head attention in transformer-based architectures \cite{Transformer, LiTransformer} to enhance explainability.
In general, attention mechanisms scale values $\bm{V} \in \mathbb{R}^{N \times d_{V}}$ based on relationships between keys $\bm{K} \in \mathbb{R}^{N \times d_{attn}}$ and queries $\bm{Q} \in \mathbb{R}^{N \times d_{attn}}$ as below:
\begin{align}
\text{Attention}(\bm{Q}, \bm{K}, \bm{V}) &= A(\bm{Q}, \bm{K}) \bm{V},
%&= \text{Softmax}(\frac{\bm{Q}\bm{K}^T}{\sqrt{d_{attn}}}) \bm{V} ,
\label{attn}
\end{align}
where $A()$ is a normalization function. A common choice is scaled dot-product attention \cite{Transformer}: 
\begin{align}
A(\bm{Q}, \bm{K}) = \text{Softmax}({\bm{Q}\bm{K}^T}/{\sqrt{d_{attn}}}).
\label{attn_normalize}
\end{align}

To improve the learning capacity of the standard attention mechanism, multi-head attention is proposed in \cite{Transformer}, employing different heads for different representation subspaces:
\begin{align}
\text{MultiHead} (\bm{Q}, \bm{K}, \bm{V}) = [\bm{H}_1, \dots,  \bm{H}_{m_H}]~\bm{W}_H, &\\
\bm{H}_h = \text{Attention}(\bm{Q}~\bm{W}_{Q}^{(h)},  \bm{K}~\bm{W}_{K}^{(h)}, \bm{V}~\bm{W}_{V}^{(h)} ), & 
\end{align}
where  $\bm{W}_{K}^{(h)} \in \mathbb{R}^{d_{model} \times d_{attn}}$, $\bm{W}_{Q}^{(h)} \in \mathbb{R}^{d_{model} \times d_{attn}}$, $\bm{W}_{V}^{(h)} \in \mathbb{R}^{d_{model} \times d_{V}}$ are head-specific weights for keys, queries and values, and $\bm{W}_H \in \mathbb{R}^{(m_H \cdot d_{V})\times d_{model}}$ linearly combines outputs concatenated from all heads $\bm{H}_h$. 

Given that different values are used in each head, attention weights alone would not be indicative of a particular feature's importance. As such, we modify multi-head attention to share values in each head, and employ additive aggregation of all heads:
\begin{align}
&\text{InterpretableMultiHead}(\bm{Q},\bm{K},\bm{V}) = \tilde{\bm{H}}~\bm{W}_H,   &
\end{align}
\begin{align}
\tilde{\bm{H}} & = \tilde{A}(\bm{Q}, \bm{K})~ \bm{V}~\bm{W}_V, &
\label{eqn:interp_multihead_attn_mat}\\
& = \bigg\{ 1/H \sum\nolimits_{h=1}^{m_H} A\left(\bm{Q}~\bm{W}_{Q}^{(h)}, \bm{K}~\bm{W}_{K}^{(h)}\right) \bigg\} \bm{V}~\bm{W}_{V},& \label{eqn:interp_multihead_ave}\\
& = 1/H \sum\nolimits_{h=1}^{m_H} \text{Attention}(\bm{Q}~\bm{W}_{Q}^{(h)}, \bm{K}~\bm{W}_{K}^{(h)}, \bm{V}~\bm{W}_{V}), &
\end{align}
where $\bm{W}_{V} \in \mathbb{R}^{d_{model} \times d_{V}}$ are value weights shared across all heads, and $\bm{W}_H \in \mathbb{R}^{d_{attn} \times d_{model}}$ is used for final linear mapping. From Eq. \eqref{eqn:interp_multihead_ave}, we see that each head can learn different temporal patterns, while attending to a common set of input features -- which can be interpreted as a simple ensemble over attention weights into combined matrix $\tilde{A}(\bm{Q}, \bm{K})$ in Eq. \eqref{eqn:interp_multihead_attn_mat}. Compared to  $A(\bm{Q}, \bm{K})$ in Eq. \eqref{attn_normalize}, $\tilde{A}(\bm{Q}, \bm{K})$ yields an increased representation capacity in an efficient way.

\subsection{Temporal Fusion Decoder}
\label{sec:temporal_fusion}
The temporal fusion decoder uses the series of layers described below to learn temporal relationships present in the dataset:

\subsubsection{Locality Enhancement with Sequence-to-Sequence Layer}
\label{sec:lstm_layer}
In time series data, points of significance are often identified in relation to their surrounding values -- such as anomalies, change-points or cyclical patterns. Leveraging local context, through the construction of features that utilize pattern information on top of point-wise values, can thus lead to performance improvements in attention-based architectures. For instance, \cite{LiTransformer} adopts a single convolutional layer for locality enhancement -- extracting local patterns using the same filter across all time. However, this might not be suitable for cases when observed inputs exist, due to the differing number of past and future inputs.
%While \cite{LiTransformer} uses convolutional filters to learn local patterns, this requires the definition of a fixed filter size, necessitating the same lookback window is used for pattern extraction at all time steps. However, determining a single optimal lookback window is challenging in practice, as it can vary significantly between time series. For instance, dynamic time warping \cite{DTW} is traditionally used to align trajectories for pattern extraction -- by stretching (or compressing) a path to improve its fit to a given template.
As such, we propose the application of a sequence-to-sequence model to naturally handle these differences -- feeding $\tilde{\bm{\xi}}_{t-k:t}$ into the encoder and $\tilde{\bm{\xi}}_{t+1:t+\tau_{max}}$ into the decoder. This then generates a set of uniform temporal features which serve as inputs into the temporal fusion decoder itself -- denoted by $\bm{\phi}(t, n) \in \left\{ \bm{\phi}(t, -k), \dots, \bm{\phi}(t, \tau_{max}) \right\}$ with $n$ being a position index. For comparability with commonly-used sequence-to-sequence baselines, we consider the use of an LSTM encoder-decoder -- although other models can potentially be adopted as well. This also serves as a replacement for standard positional encoding, providing an appropriate inductive bias for the time ordering of the inputs. Moreover, to allow static metadata to influence local processing, we use the $\bm{c}_{c}, \bm{c}_{h}$ context vectors from the static covariate encoders to initialize the cell state and hidden state respectively for the first LSTM in the layer. We also employ a gated skip connection over this layer:
\begin{align}
\tilde{\bm{\phi}}(t, n) = \text{LayerNorm}\left( \tilde{\bm{\xi}}_{t+n} + \text{GLU}_{\tilde{\phi}} (\bm{\phi}(t, n)) \right) , 
\end{align} 
where $n \in [-k, \tau_{max}]$ is a position index.

\subsubsection{Static Enrichment Layer}
\label{sec:static_enrichment}
As static covariates often have a significant influence on the temporal dynamics (e.g. genetic information on disease risk), we introduce a static enrichment layer that enhances temporal features with static metadata. For a given position index $n$, static enrichment takes the form:
\begin{align}
\bm{\theta}(t, n) = \text{GRN}_{\theta} \left( \tilde{\bm{\phi}}(t, n), \bm{c}_e \right), &
\end{align}
where the weights of $\text{GRN}_{\phi}$ are shared across the entire layer, and $\bm{c}_e$ is a context vector from a static covariate encoder.
\subsubsection{Temporal Self-Attention Layer} 
Following static enrichment, we next apply self-attention. All static-enriched temporal features are first grouped into a single matrix -- i.e. $\bm{\Theta}(t) = [\bm{\theta}(t, -k), \dots,$ $\bm{\theta}(t, \tau)]^T$ -- and interpretable multi-head attention (see Sec. \ref{sec:interp_multi_head}) is applied at each forecast time (with $N=\tau_{max} + k + 1$):
\begin{equation}
\bm{B}(t) = \text{InterpretableMultiHead}(\bm{\Theta}(t),\bm{\Theta}(t),\bm{\Theta}(t)),
\label{eqn:self_attn_applied}
\end{equation}
to yield $\bm{B}(t) = \left[\bm{\beta}(t, -k), \dots, \bm{\beta}(t,\tau_{max}) \right]$.
$d_{V} = d_{attn} = d_{model}/m_H$ are chosen, where $m_H$ is the number of heads.
Decoder masking \cite{Transformer, LiTransformer} is applied to the multi-head attention layer to ensure that each temporal dimension can only attend to features preceding it. Besides preserving causal information flow via masking, the self-attention layer allows TFT to pick up long-range dependencies that may be challenging for RNN-based architectures to learn. Following the self-attention layer, an additional gating layer is also applied to facilitate training:
\begin{align}
\bm{\delta}(t, n) = \text{LayerNorm}(\bm{\theta}(t, n) + \text{GLU}_\delta(\bm{\beta}(t, n))).
\end{align}

\subsubsection{Position-wise Feed-forward Layer} 
\label{sec:tranformer_positionwise_ff}
We apply an additional non-linear processing to the outputs of the self-attention layer. Similar to the static enrichment layer, this makes use of GRNs:
\begin{align}
\bm{\psi}(t, n) = \text{GRN}_{\psi}\left(\bm{\delta}(t, n)\right),
\end{align}
where the weights of $\text{GRN}_{\psi}$ are shared across the entire layer. As per Fig. \ref{fig:tft_architecture}, we also apply a gated residual connection which skips over the entire transformer block, providing a direct path to the sequence-to-sequence layer -- yielding a simpler model if additional complexity is not required, as shown below:
\begin{align}
\tilde{\bm{\psi}}(t, n) = \text{LayerNorm}\left( \tilde{\bm{\phi}}(t,n) + \text{GLU}_{\tilde{\psi}} (\bm{\psi}(t, n)) \right), 
\end{align} 

\subsection{Quantile Outputs}
In line with previous work \cite{MQRNN}, TFT also generates prediction intervals on top of point forecasts. This is achieved by the simultaneous prediction of various percentiles (e.g. $10^{th}$, $50^{th}$ and $90^{th}$) at each time step. Quantile forecasts are generated using linear transformation of the output from the temporal fusion decoder:
\begin{equation}
\hat{y}(q, t, \tau) = \bm{W}_{q}~\tilde{\bm{\psi}}(t, \tau)  + b_{q},
\end{equation}
where $\bm{W}_{q} \in \mathbb{R}^{1 \times d}, b_{q} \in \mathbb{R}$ are linear coefficients for the specified quantile $q$. We note that forecasts are only generated for horizons in the future -- i.e. $\tau \in \{1, \dots, \tau_{max} \}$.

\section{Loss Functions}
TFT is trained by jointly minimizing the quantile loss \cite{MQRNN}, summed across all quantile outputs:
\begin{equation}
\mathcal{L}(\Omega, \bm{W}) = \sum\nolimits_{y_t \in \Omega} \sum\nolimits_{q \in \mathcal{Q}} \sum\nolimits_{\tau=1}^{\tau_{max}} \frac{ QL\left (y_{t},~\hat{y}(q, t-\tau, \tau),~q \right)}{M \tau_{max}}
\end{equation}
\begin{equation}
QL(y, ~\hat{y},~q) = q (y - \hat{y})_+ + (1-q) ( \hat{y} - y)_+ , 
\end{equation}
where $\Omega$ is the domain of training data containing $M$ samples, $\bm{W}$ represents the weights of TFT, $\mathcal{Q}$  is the set of output quantiles (we use $\mathcal{Q} = \{ 0.1, 0.5, 0.9 \}$ in our experiments, and $(.)_+ = \text{max}(0, .)$. For out-of-sample testing, we evaluate the normalized quantile losses across the entire forecasting horizon -- focusing on P50 and P90 risk for consistency with previous work \cite{DeepAR, DSSM, LiTransformer}:
\begin{equation}
q\text{-Risk} = \frac{2 \sum_{y_t \in \tilde{\Omega}} \sum_{\tau=1}^{\tau_{max}}  QL\left (y_{t},~\hat{y}(q, t-\tau, \tau),~q \right)}{ \sum_{y_t \in \tilde{\Omega}} \sum_{\tau=1}^{\tau_{max}} |y_t|},
\end{equation}
where $\tilde{\Omega}$ is the domain of test samples. Full details on hyperparameter optimization and training can be found in \ref{apdx:dataset_info}.
 
\section{Performance Evaluation}
\subsection{Datasets} 
\label{sec:data_overview}
We choose datasets to reflect commonly observed characteristics across a wide range of challenging multi-horizon forecasting problems.
To establish a baseline and position with respect to prior academic work, we first evaluate performance on the Electricity and Traffic datasets used in \cite{DeepAR,DSSM,LiTransformer} -- which focus on simpler univariate time series containing known inputs only alongside the target. Next, the Retail dataset helps us benchmark the model using the full range of complex inputs observed in multi-horizon prediction applications (see Sec. \ref{sec:problem_def}) -- including rich static metadata and observed time-varying inputs. Finally, to evaluate robustness to over-fitting on smaller noisy datasets, we consider the financial application of volatility forecasting -- using a dataset much smaller than others. Broad descriptions of each dataset can be found below: 
\begin{itemize}[noitemsep, nolistsep, leftmargin=*]
\item \textbf{Electricity:} The UCI Electricity Load Diagrams Dataset, containing the electricity consumption of 370 customers -- aggregated on an hourly level as in \cite{MatrixFactorisation}. In accordance with \cite{DeepAR}, we use the past week (i.e. 168 hours) to forecast over the next 24 hours. 
\item \textbf{Traffic:} The UCI PEM-SF Traffic Dataset describes the occupancy rate (with $y_t \in [0,1]$) of 440 SF Bay Area freeways -- as in \cite{MatrixFactorisation}. It is also aggregated on an hourly level as per the electricity dataset, with the same look back window and forecast horizon.
\item \textbf{Retail:} Favorita Grocery Sales Dataset from the Kaggle competition \cite{favorita}, that combines metadata for different products and the stores, along with other exogenous time-varying inputs sampled at the daily level. We forecast log product sales 30 days into the future, using 90 days of past information.
\item \textbf{Volatility (or Vol.):} The OMI realized library \cite{omi} contains daily realized volatility values of 31 stock indices computed from intraday data, along with their daily returns. For our experiments, we consider forecasts over the next week (i.e. 5 business days) using information over the past year (i.e. 252 business days).
\end{itemize}

\subsection{Training Procedure}
For each dataset, we partition all time series into 3 parts -- a training set for learning, a validation set for hyperparameter tuning, and a hold-out test set for performance evaluation. 
Hyperparameter optimization is conducted via random search, using 240 iterations for Volatility, and 60 iterations for others.
Full search ranges for all hyperparameters are below, with datasets and optimal model parameters listed in Table \ref{tab:hyperparams}. 

\begin{itemize}[noitemsep]
    \item \textbf{State size} -- 10, 20, 40, 80, 160, 240, 320
    \item \textbf{Dropout rate} -- 0.1, 0.2, 0.3, 0.4, 0.5, 0.7, 0.9
    \item \textbf{Minibatch size} -- 64, 128, 256
    \item \textbf{Learning rate} -- 0.0001, 0.001, 0.01
    \item \textbf{Max. gradient norm} -- 0.01, 1.0, 100.0
    \item \textbf{Num. heads} -- 1, 4
\end{itemize}

To preserve explainability, we adopt only a single interpretable multi-head attention layer. For ConvTrans \cite{LiTransformer}, we use the same fixed stack size (3 layers) and number of heads (8 heads) as in \cite{LiTransformer}. We keep the same attention model, and treat kernel sizes for the convolutional processing layer as a hyperparameter ($\in \{1, 3 ,6, 9\}$) -- as optimal kernel sizes are observed to be dataset dependent \cite{LiTransformer}. An open-source implementation of the TFT on these datasets can be found on GitHub\footnote{\ifthenelse{\boolean{anon}}{Made available at publication time.}{URL: \url{https://github.com/google-research/google-research/tree/master/tft}}} for full reproducibility.

\subsection{Computational Cost}
Across all datasets, each TFT model was also trained on a single GPU, and can be deployed without the need for extensive computing resources. For instance, using a NVIDIA Tesla V100 GPU, our optimal TFT model (for Electricity dataset) takes just slightly over 6 hours to train (each epoch being roughly 52 mins). The batched inference on the entire validation dataset (consisting 50,000 samples) takes 8 minutes. TFT training and inference times can be further reduced with hardware-specific optimizations.

\begin{table}[th]
\caption{Information on dataset and optimal TFT configuration.}
\centerline{ \begin{tabular}{@{}l|llll@{}}
\toprule
\textbf{}                    & \textbf{Electricity} & \textbf{Traffic} & \textbf{Retail} & \textbf{Vol.} \\ \midrule
{\ul \textbf{Dataset Details}}       &                      &                  &                 &                     \\
Target Type                  & $\mathbb{R}$         & [0, 1]   & $\mathbb{R}$    & $\mathbb{R}$        \\
Number of Entities                  & 370                  & 440              & 130k            & 41                  \\
Number of Samples                   & 500k                 & 500k             & 500k            & $\sim$100k          \\ \midrule
{\ul \textbf{Network Parameters}} &                      &                  &                 &                     \\
$k$                         & 168                  & 168              & 90              & 252                 \\
$\tau_{max}$                & 24                   & 24               & 30              & 5                   \\
Dropout Rate                 & 0.1                  & 0.3              & 0.1             & 0.3                 \\
State Size                   & 160                  & 320              & 240             & 160                 \\
Number of Heads                     & 4                    & 4                & 4               & 1                   \\ \midrule
{\ul \textbf{Training Parameters}} &                      &                  &                 &                     \\
Minibatch Size                                               & 64                   & 128              & 128             & 64                  \\
Learning Rate                                                & 0.001                & 0.001            & 0.001           & 0.01                \\
Max Gradient Norm & 0.01                 & 100              & 100             & 0.01                \\ \bottomrule
\end{tabular}  }
\label{tab:hyperparams}
\end{table}

\subsection{Benchmarks}
\begin{table}[p]
\caption{P50 and P90 quantile losses on a range of real-world datasets. Percentages in brackets reflect the increase in quantile loss versus TFT (lower $q$-Risk better), with TFT outperforming competing methods across all experiments, improving on the next best alternative method (\underline{underlined}) between 3\% and 26\%.}
\begin{subtable}[t]{1.0\textwidth}
\centerline{\begin{tabular}{@{}llllll@{}}
\toprule
{\color[HTML]{000000} }                     & {\color[HTML]{000000} \textbf{ARIMA}}     & {\color[HTML]{000000} \textbf{ETS}}     & {\color[HTML]{000000} \textbf{TRMF}}  & {\color[HTML]{000000} \textbf{DeepAR}} & {\color[HTML]{000000} \textbf{DSSM}} \\ \midrule
{\color[HTML]{000000} \textbf{Electricity}} & {\color[HTML]{000000} 0.154 (\textit{+180\%})}      & {\color[HTML]{000000} 0.102 (\textit{+85\%})}     & {\color[HTML]{000000} 0.084 (\textit{+53\%})}   & {\color[HTML]{000000} 0.075 (\textit{+36\%})}    & {\color[HTML]{000000} 0.083 (\textit{+51\%})}  \\
{\color[HTML]{000000} \textbf{Traffic}}     & {\color[HTML]{000000} 0.223 (\textit{+135\%})}      & {\color[HTML]{000000} 0.236 (\textit{+148\%})}    & {\color[HTML]{000000} 0.186 (\textit{+96\%})}   & {\color[HTML]{000000} 0.161 (\textit{+69\%})}    & {\color[HTML]{000000} 0.167 (\textit{+76\%})}  \\ \midrule
{\color[HTML]{000000} }                     & {\color[HTML]{000000} \textbf{ConvTrans}} & {\color[HTML]{000000} \textbf{Seq2Seq}} & {\color[HTML]{000000} \textbf{MQRNN}} & {\color[HTML]{000000} \textbf{TFT}}    & {\color[HTML]{000000} \textbf{}}     \\ \midrule
{\color[HTML]{000000} \textbf{Electricity}} & {\color[HTML]{000000} 0.059 (\ul{\textit{+7\%}})}        & {\color[HTML]{000000} 0.067 (\textit{+22\%})}     & {\color[HTML]{000000} 0.077 (\textit{+40\%})}   & {\color[HTML]{000000} \textbf{0.055*}} & {\color[HTML]{000000} }              \\
{\color[HTML]{000000} \textbf{Traffic}}     & {\color[HTML]{000000} 0.122 (\textit{+28\%})}       & {\color[HTML]{000000} 0.105 (\ul{\textit{+11\%}})}     & {\color[HTML]{000000} 0.117 (\textit{+23\%})}   & {\color[HTML]{000000} \textbf{0.095*}} & {\color[HTML]{000000} }              \\ \bottomrule
\end{tabular}}
\caption{P50 losses on simpler univariate datasets.}
\end{subtable}
\begin{subtable}[t]{1.0\textwidth}
\centerline{\begin{tabular}{@{}llllll@{}}
\toprule
{\color[HTML]{000000} }                     & {\color[HTML]{000000} \textbf{ARIMA}}     & {\color[HTML]{000000} \textbf{ETS}}     & {\color[HTML]{000000} \textbf{TRMF}}  & {\color[HTML]{000000} \textbf{DeepAR}} & {\color[HTML]{000000} \textbf{DSSM}} \\ \midrule
{\color[HTML]{000000} \textbf{Electricity}} & {\color[HTML]{000000} 0.102 (\textit{+278\%})}      & {\color[HTML]{000000} 0.077 (\textit{+185\%})}    & {\color[HTML]{000000} -}              & {\color[HTML]{000000} 0.040 (\textit{+48\%})}    & {\color[HTML]{000000} 0.056 (\textit{+107\%})} \\
{\color[HTML]{000000} \textbf{Traffic}}     & {\color[HTML]{000000} 0.137 (\textit{+94\%})}       & {\color[HTML]{000000} 0.148 (\textit{+110\%})}    & {\color[HTML]{000000} -}              & {\color[HTML]{000000} 0.099 (\textit{+40\%})}    & {\color[HTML]{000000} 0.113 (\textit{+60\%})}  \\ \midrule
{\color[HTML]{000000} }                     & {\color[HTML]{000000} \textbf{ConvTrans}} & {\color[HTML]{000000} \textbf{Seq2Seq}} & {\color[HTML]{000000} \textbf{MQRNN}} & {\color[HTML]{000000} \textbf{TFT}}    & {\color[HTML]{000000} \textbf{}}     \\ \midrule
{\color[HTML]{000000} \textbf{Electricity}} & {\color[HTML]{000000} 0.034 (\ul{\textit{+26\%}})}       & {\color[HTML]{000000} 0.036 (\textit{+33\%})}     & {\color[HTML]{000000} 0.036 (\textit{+33\%})}   & {\color[HTML]{000000} \textbf{0.027*}} & {\color[HTML]{000000} }              \\
{\color[HTML]{000000} \textbf{Traffic}}     & {\color[HTML]{000000} 0.081 (\textit{+15\%})}       & {\color[HTML]{000000} 0.075 (\ul{\textit{+6\%}})}      & {\color[HTML]{000000} 0.082 (\textit{+16\%})}   & {\color[HTML]{000000} \textbf{0.070*}} & {\color[HTML]{000000} }              \\ \bottomrule
\end{tabular}}
\caption{P90 losses on simpler univariate datasets.}
\end{subtable}
\vfill
\begin{subtable}[t]{\textwidth}
\centerline{\begin{tabular}{@{}llllll@{}}
\toprule
{\color[HTML]{000000} }                & {\color[HTML]{000000} \textbf{DeepAR}} & {\color[HTML]{000000} \textbf{CovTrans}} & {\color[HTML]{000000} \textbf{Seq2Seq}} & {\color[HTML]{000000} \textbf{MQRNN}} & {\color[HTML]{000000} \textbf{TFT}}   \\ \midrule
{\color[HTML]{000000} \textbf{Vol.}}   & {\color[HTML]{000000} 0.050 (\textit{+28\%})}   & {\color[HTML]{000000} 0.047 (\textit{+20\%})}     & {\color[HTML]{000000} 0.042 (\ul{\textit{+7\%}})}     & {\color[HTML]{000000} 0.042 (\ul{\textit{+7\%}})}   & {\color[HTML]{000000} \textbf{0.039*}} \\
{\color[HTML]{000000} \textbf{Retail}} & {\color[HTML]{000000} 0.574 (\textit{+62\%})}   & {\color[HTML]{000000} 0.429 (\textit{+21\%})}     & {\color[HTML]{000000} 0.411 (\textit{+16\%})}    & {\color[HTML]{000000} 0.379 (\ul{\textit{+7\%}})}   & {\color[HTML]{000000} \textbf{0.354*}} \\ \bottomrule
\end{tabular}}
\caption{P50 losses on datasets with rich static or observed inputs.}
\end{subtable}
\begin{subtable}[t]{\textwidth}
\centerline{\begin{tabular}{@{}llllll@{}}
\toprule
{\color[HTML]{000000} }                & {\color[HTML]{000000} \textbf{DeepAR}} & {\color[HTML]{000000} \textbf{CovTrans}} & {\color[HTML]{000000} \textbf{Seq2Seq}} & {\color[HTML]{000000} \textbf{MQRNN}} & {\color[HTML]{000000} \textbf{TFT}}   \\ \midrule
{\color[HTML]{000000} \textbf{Vol.}}   & {\color[HTML]{000000} 0.024 (\textit{+21\%})}   & {\color[HTML]{000000} 0.024 (\textit{+22\%})}     & {\color[HTML]{000000} 0.021 (\ul{\textit{+8\%}})}     & {\color[HTML]{000000} 0.021 (\textit{+9\%})}   & {\color[HTML]{000000} \textbf{0.020*}}  \\
{\color[HTML]{000000} \textbf{Retail}} & {\color[HTML]{000000} 0.230 (\textit{+56\%})}   & {\color[HTML]{000000} 0.192 (\textit{+30\%})}     & {\color[HTML]{000000} 0.157 (\textit{+7\%})}     & {\color[HTML]{000000} 0.152 (\ul{\textit{+3\%}})}   & {\color[HTML]{000000} \textbf{0.147*}} \\ \bottomrule
\end{tabular}}
\caption{P90 losses on datasets with rich static or observed inputs.}
\end{subtable}
\label{tab:results}
\end{table}

We extensively compare TFT to a wide range of models for multi-horizon forecasting, based on the categories described in Sec. \ref{sec:related_works}. Hyperparameter optimization is conducted using random search over a pre-defined search space, using the same number of iterations across all benchmarks for a give dataset. Additional details are included in \ref{apdx:dataset_info}.

\textbf{Direct methods:} As TFT falls within this class of multi-horizon models, we primarily focus comparisons on deep learning models which directly generate prediction at future horizons, including: 1) simple sequence-to-sequence models with global contexts (Seq2Seq), and 2) the Multi-horizon Quantile Recurrent Forecaster (MQRNN) \cite{MQRNN}.

\textbf{Iterative methods:} To position with respect to the rich body of work on iterative models, we evaluate TFT using the same setup as \cite{DeepAR} for the Electricity and Traffic datasets. This extends the results from \cite{LiTransformer} for 1) DeepAR \cite{DeepAR}, 2) DSSM \cite{DSSM}, and 3) the Transformer-based architecture of \cite{LiTransformer} with local convolutional processing  -- which refer to as ConvTrans. For more complex datasets, we focus on the ConvTrans model given its strong outperformance over other iterative models in prior work, and DeepAR due to its popularity among practitioners. As models in this category require knowledge of all inputs in the future to generate predictions, we accommodate this for complex datasets by imputing unknown inputs with their last available value.

For simpler univariate datasets, we note that the results for ARIMA, ETS, TRMF, DeepAR, DSSM and ConvTrans have been reproduced from \cite{LiTransformer} in Table \ref{tab:results} for consistency.
 
\subsection{Results and Discussion}
Table \ref{tab:results} shows that TFT significantly outperforms all benchmarks over the variety of datasets described in Sec. \ref{sec:data_overview}. For median forecasts, TFT yields $7\%$ lower P50 and $9\%$ lower P90 losses on average compared to the next best model -- demonstrating the benefits of explicitly aligning the architecture with the general multi-horizon forecasting problem.

Comparing direct and iterative models, we observe the importance of accounting for the observed inputs -- noting the poorer results of ConvTrans on complex datasets where observed input imputation is required (i.e. Volatility and Retail). Furthermore, the benefits of quantile regression are also observed when targets are not captured well by Gaussian distributions with direct models outperforming in those scenarios. This can be seen, for example, from the Traffic dataset where target distribution is significantly skewed -- with more than $90\%$ of occupancy rates falling between 0 and 0.1, and the remainder distributed evenly until 1.0.

\subsection{Ablation Analysis}
\label{sec:ablation}
To quantify the benefits of each of our proposed architectural contribution, we perform an extensive ablation analysis -- removing each component from the network as below, and quantifying the percentage increase in loss versus the original architecture:  
\begin{itemize}[noitemsep, nolistsep, leftmargin=*]
\item \textbf{Gating layers:} We ablate by replacing each GLU layer (Eq. \eqref{eqn:component_gate}) with a simple linear layer followed by ELU.
\item \textbf{Static covariate encoders:} We ablate by setting all context vectors to zero -- i.e. $\bm{c}_s{=}\bm{c}_e{=}\bm{c}_c{=}\bm{c}_h{=}\bm{0}$ -- and concatenating all transformed static inputs to all time-dependent past and future inputs.
\item \textbf{Instance-wise variable selection networks:} We ablate by replacing the softmax outputs of Eq. \ref{eq:var_selection} with trainable coefficients, and removing the networks generating the variable selection weights. We retain, however, the variable-wise GRNs (see Eq. \eqref{sec:varselect_grn}), maintaining a similar amount of non-linear processing.
\item \textbf{Self-attention layers:}  We ablate by replacing the attention matrix of the interpretable multi-head attention layer (Eq. \ref{eqn:interp_multihead_attn_mat}) with a matrix of trainable parameters $\bm{W}_A$ -- i.e. $\tilde{A}(\bm{Q}, \bm{K}) = \bm{W}_A$, where $\bm{W}_A \in \mathbb{R}^{N \times N}$. This prevents TFT from attending to different input features at different times, helping evaluation of the importance of instance-wise attention weights. 
\item \textbf{Sequence-to-sequence layers for local processing:} We ablate by replacing the sequence-to-sequence layer of Sec. \ref{sec:lstm_layer} with standard positional encoding used in \cite{Transformer}. 
\end{itemize}

\begin{figure}[t]
\begin{subfigure}{1.0\textwidth}
\centering
    \includegraphics[width=1.0\linewidth]{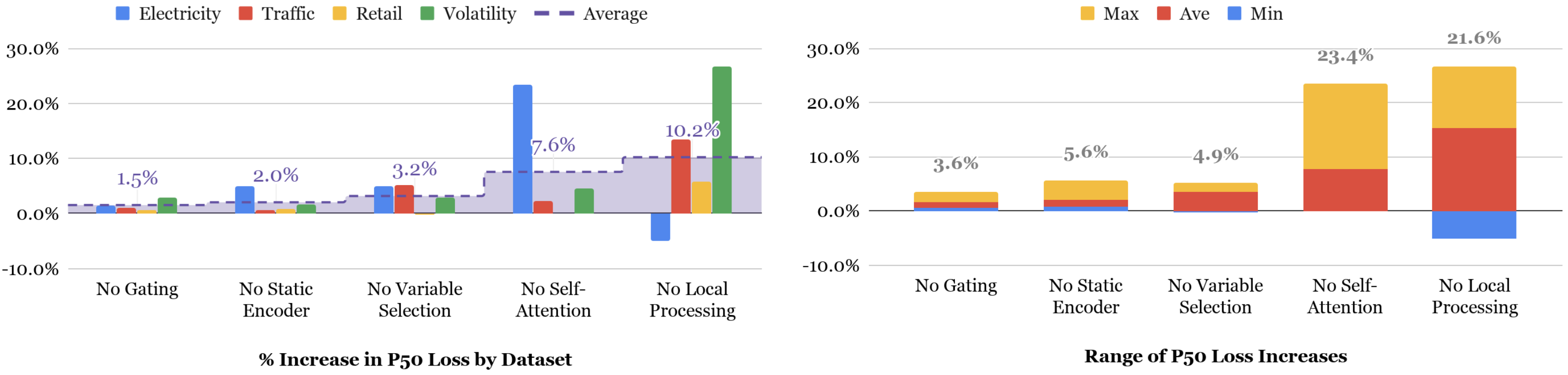}
    \caption{Changes in P50 losses across ablation tests}
\end{subfigure} \\
\begin{subfigure}{1.0\textwidth}
\centering
    \includegraphics[width=1.0\linewidth]{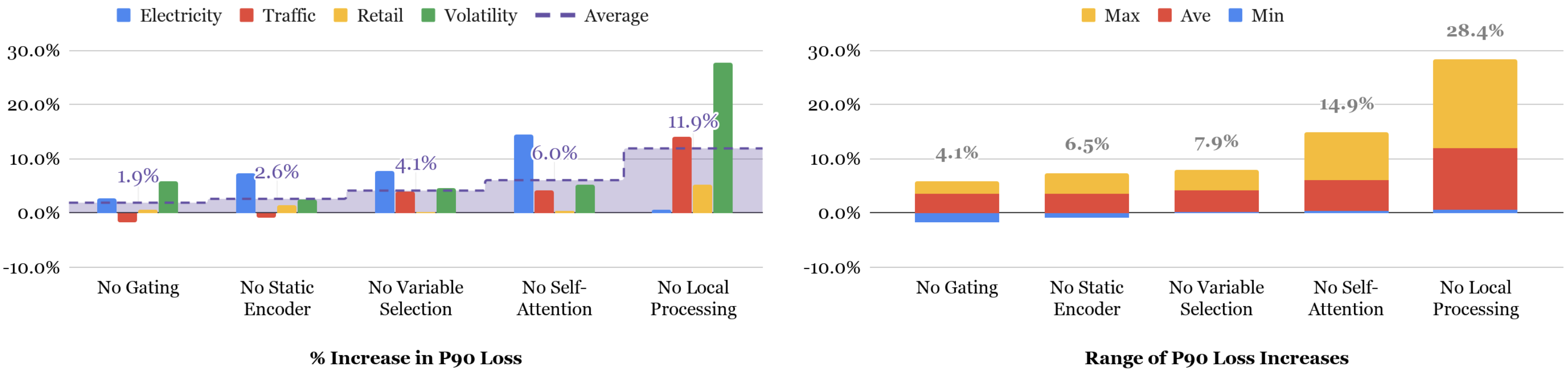}
    \caption{Changes in P90 losses across ablation tests}
\end{subfigure} 
\caption{Results of ablation analysis. Both a) and b) show the impact of ablation on the P50 and P90 losses respectively. Results per dataset shown on the left, and the range across datasets shown on the right. While the precise importance of each is dataset-specific, all components contribute significantly on the whole -- with the maximum percentage increase over all datasets ranging from $3.6\%$ to $23.4\%$ for P50 losses, and similarly from $4.1\%$ to $28.4\%$ for P90 losses.}
\label{fig:ablation}
\end{figure}

Ablated networks are trained across for each dataset using the hyperparameters of Table \ref{tab:hyperparams}. Fig. \ref{fig:ablation} shows that the effects on both P50 and P90 losses are similar across all datasets, with all components contributing to performance improvements on the whole. 

In general, the components responsible for capturing temporal relationships, local processing and self-attention layers, have the largest impact on performance, with P90 loss increases of $>6\%$ on average and $> 20\%$ on select datasets when ablated. 
The diversity across time series datasets can also be seen from the differences in the ablation impact of the respective temporal components. Concretely, while local processing is critical in Traffic, Retail and Volatility, lower post-ablation P50 losses indicate that it can be detrimental in Electricity -- with the self-attention layer playing a more vital role. A possible explanation is that persistent daily seasonality appears to dominate other temporal relationships in the Electricity dataset. 
For this dataset, Table \ref{tab:apdx_varselect} of \ref{apdx:interp} also shows that the hour-of-day has the largest variable importance score across all temporal inputs, exceeding even the target (i.e. Power Usage) itself. In contrast to other dataset where past target observations are more significant (e.g. Traffic), direct attention to previous days seem to help learning daily seasonal patterns in Electricity -- with local processing between adjacent time steps being less necessary. We can account for this by treating the sequence-to-sequence architecture in the temporal fusion decoder as a hyperparameter to tune, including an option for simple positional encoding without any local processing.

Static covariate encoders and instance-wise variable selection have the next largest impact -- increasing P90 losses by more than $2.6\%$ and $4.1\%$ on average. The biggest benefits of these are observed for electricity dataset, where some of the input features get very low importance.

Finally, gating layer ablation also shows increases in P90 losses, with a $1.9\%$ increase on average. This is the most significant on the volatility (with a $4.1\%$ P90 loss increase), underlying the benefit of component gating for smaller and noisier datasets. 

\section{Interpretability Use Cases}
\label{sec:interp}
Having established the performance benefits of our model, we next demonstrate how our model design allows for analysis of its individual components to interpret the general relationships it has learned. 
We demonstrate three interpretability use cases: 
(1) examining the importance of each input variable in prediction, 
(2) visualizing persistent temporal patterns, and 
(3) identifying any regimes or events that lead to significant changes in temporal dynamics. 
In contrast to other examples of attention-based interpretability \cite{AttendAndDiagnose,LiTransformer,AttentionDiseaseProgression} which zoom in on interesting but instance-specific examples, our methods focus on ways to aggregate the patterns across the entire dataset -- extracting generalizable insights about temporal dynamics.

\subsection{Analyzing Variable Importance}

\begin{table}[p]
\caption{Variable importance for the Retail dataset. The $10^{th}$, $50^{th}$ and $90^{th}$ percentiles of the variable selection weights are shown, with values larger than 0.1 highlighted in purple. For static covariates, the largest weights are attributed to variables which uniquely identify different entities (i.e. item number and store number). For past inputs, past values of the target (i.e. log sales) are critical as expected, as forecasts are extrapolations of past observations. For future inputs, promotion periods and national holidays have the greatest influence on sales forecasts, in line with periods of increased customer spending. }
\label{tab:variable_importance}
\centerline{
\begin{subtable}[t]{0.45\textwidth}
\hspace{-0.5cm}
\begin{tabular}[t]{@{}llll@{}}
\toprule
                          & \textbf{10\%} & \textbf{50\%} & \textbf{90\%} \\ \midrule
\textbf{Item Num}         & \textcolor{violet}{\textbf{0.198}}         & \textcolor{violet}{\textbf{0.230}}         & \textcolor{violet}{\textbf{0.251}}         \\
\textbf{Store Num}        & \textcolor{violet}{\textbf{0.152}}         & \textcolor{violet}{\textbf{0.161}}         & \textcolor{violet}{\textbf{0.170}}         \\
\textbf{City}             & 0.094         & 0.100         & 0.124         \\
\textbf{State}            & 0.049         & 0.060         & 0.083         \\
\textbf{Type}             & 0.005         & 0.006         & 0.008         \\
\textbf{Cluster}          & \textcolor{violet}{\textbf{0.108}}         & \textcolor{violet}{\textbf{0.122}}         & \textcolor{violet}{\textbf{0.133}}         \\
\textbf{Family}           & 0.063         & 0.075         & 0.079         \\
\textbf{Class}            & \textcolor{violet}{\textbf{0.148}}         & \textcolor{violet}{\textbf{0.156}}         & \textcolor{violet}{\textbf{0.163}}         \\
\textbf{Perishable}       & 0.084         & 0.085         & 0.088         \\ 
& & & \\
& & & \\
\bottomrule
\end{tabular}
\caption{Static Covariates}
\end{subtable}\hfill
\begin{subtable}[t]{0.45\textwidth}
\hspace{-0.5cm}
\begin{tabular}[t]{@{}llll@{}}
\toprule
                          & \textbf{10\%} & \textbf{50\%} & \textbf{90\%} \\ \midrule
\textbf{Transactions}     & 0.029         & 0.033         & 0.037         \\
\textbf{Oil}              & 0.062         & 0.081         & 0.105         \\
\textbf{On-promotion}     & 0.072         & 0.075         & 0.078         \\
\textbf{Day of Week}      & 0.007         & 0.007         & 0.008         \\
\textbf{Day of Month}     & 0.083         & 0.089         & 0.096         \\
\textbf{Month}            & \textcolor{violet}{\textbf{0.109}}         & \textcolor{violet}{\textbf{0.122}}         & \textcolor{violet}{\textbf{0.136}}         \\
\textbf{National Hol}     & \textcolor{violet}{\textbf{0.131}}         & \textcolor{violet}{\textbf{0.138}}         & \textcolor{violet}{\textbf{0.145}}         \\
\textbf{Regional Hol}     & 0.011         & 0.014         & 0.018         \\
\textbf{Local Hol}        & 0.056         & 0.068         & 0.072         \\
\textbf{Open}             & 0.027         & 0.044         & 0.067         \\
\textbf{Log Sales}        & \textcolor{violet}{\textbf{0.304}}         & \textcolor{violet}{\textbf{0.324}}         & \textcolor{violet}{\textbf{0.353}}         \\\bottomrule
\end{tabular}
\caption{Past Inputs}
\end{subtable}}
\vfill
\centering
\begin{subtable}[t]{0.45\textwidth}
\begin{tabular}[t]{@{}llll@{}}
\toprule
                          & \textbf{10\%} & \textbf{50\%} & \textbf{90\%} \\ \midrule
\textbf{On-promotion}     & \textcolor{violet}{\textbf{0.155}}         & \textcolor{violet}{\textbf{0.170}}         & \textcolor{violet}{\textbf{0.182}}         \\
\textbf{Day of Week}      & 0.029         & 0.065         & 0.089         \\
\textbf{Day of Month}     & \textcolor{violet}{\textbf{0.056}}         & \textcolor{violet}{\textbf{0.116}}         & \textcolor{violet}{\textbf{0.138}}         \\
\textbf{Month}            & \textcolor{violet}{\textbf{0.111}}         & \textcolor{violet}{\textbf{0.155}}         & \textcolor{violet}{\textbf{0.240}}         \\
\textbf{National Hol}     & \textcolor{violet}{\textbf{0.145}}         & \textcolor{violet}{\textbf{0.220}}         & \textcolor{violet}{\textbf{0.242}}         \\
\textbf{Regional Hol}     & 0.012         & 0.014         & 0.060         \\
\textbf{Local Hol}        & \textcolor{violet}{\textbf{0.116}}         & \textcolor{violet}{\textbf{0.151}}         & \textcolor{violet}{\textbf{0.239}}         \\
\textbf{Open}             & 0.088         & 0.095         & 0.097         \\ 
\bottomrule
\end{tabular}
\caption{Future Inputs}
\end{subtable}
\end{table}

We first quantify variable importance by analyzing the variable selection weights described in Sec. \ref{sec:variable_selection}. Concretely, we aggregate selection weights (i.e. $v_{\chi_t}^{(j)}$ in Eq. \eqref{eq:var_selection_sum}) for each variable across our entire test set, recording the $10^{th}$, $50^{th}$ and $90^{th}$ percentiles of each sampling distribution. As the Retail dataset contains the full set of available input types (i.e. static metadata, known inputs, observed inputs and the target), we present the results for its variable importance analysis in Table \ref{tab:variable_importance}. We also note similar findings in other datasets, which are documented in \ref{apdx:interp_variable} for completeness. On the whole, the results show that the TFT extracts only a subset of key inputs that intuitively play a significant role in predictions.
The analysis of persistent temporal patterns is often key to understanding the time-dependent relationships present in a given dataset. For instance, lag models are frequently adopted to study length of time required for an intervention to take effect \cite{CausalLag2} -- such as the impact of a government's increase in public expenditure on the resultant growth in Gross National Product \cite{LagModels}. Seasonality models are also commonly used in econometrics to identify periodic patterns in a target-of-interest \cite{Seasonality} and measure the length of each cycle. From a practical standpoint, model builders can use these insights to further improve the forecasting model -- for instance by increasing the receptive field to incorporate more history if attention peaks are observed at the start of the lookback window, or by engineering features to directly incorporate seasonal effects. 
As such, using the attention weights present in the self-attention layer of the temporal fusion decoder, we present a method to identify similar persistent patterns -- by measuring the contributions of features at fixed lags in the past on forecasts at various horizons. Combining Eq. \eqref{eqn:interp_multihead_attn_mat} and \eqref{eqn:self_attn_applied}, we see that the self-attention layer contains a matrix of attention weights at each forecast time $t$ -- i.e. $\tilde{A}(\bm{\phi}(t), \bm{\phi}(t))$. 
%
%\begin{equation}
%%\tilde{A}(\bm{\phi}(t), \bm{\phi}(t)) =  \begin{bmatrix}
%\alpha(t, -k, -k) & \dots & \alpha(t, \tau_{max}, -k) \\
%\vdots,  & \ddots & \vdots \\
%\alpha(t, -k, \tau_{max}) & \dots & \alpha(t, \tau_{max}, \tau_{max})\\
%\end{bmatrix},
%\end{equation}
Multi-head attention outputs at each forecast horizon $\tau$ $\left( \text{i.e. } \bm{\beta}(t, \tau) \right)$ can then be described as an attention-weighted sum of lower level features at each position $n$:
\begin{align}
\bm{\beta}(t, \tau) = \sum\nolimits_{n=-k}^{\tau_{max}} \alpha(t, n, \tau) ~ \tilde{\bm{\theta}}(t, n),
\end{align}
where $\alpha(t, n, \tau)$ is the $(\tau, n)$-th element of $\tilde{A}(\bm{\phi}(t), \bm{\phi}(t))$, and $\tilde{\bm{\theta}}(t, n)$ is a row of $\tilde{\bm{\Theta}}(t) = \bm{\Theta}(t) \bm{W}_V$. Due to decoder masking, we also note that $\alpha(t, i, j) = 0$, $\forall i>j$. For each forecast horizon $\tau$, the importance of a previous time point $n <\tau$ can hence be determined by analyzing distributions of $\alpha(t, n, \tau)$ across all time steps and entities. 

\subsection{Visualizing Persistent Temporal Patterns}
\label{sec:interp_persistent_patterns}
\begin{figure}[pbt]
\centering
\begin{subfigure}{0.49\textwidth}
\centering
    \includegraphics[width=\linewidth]{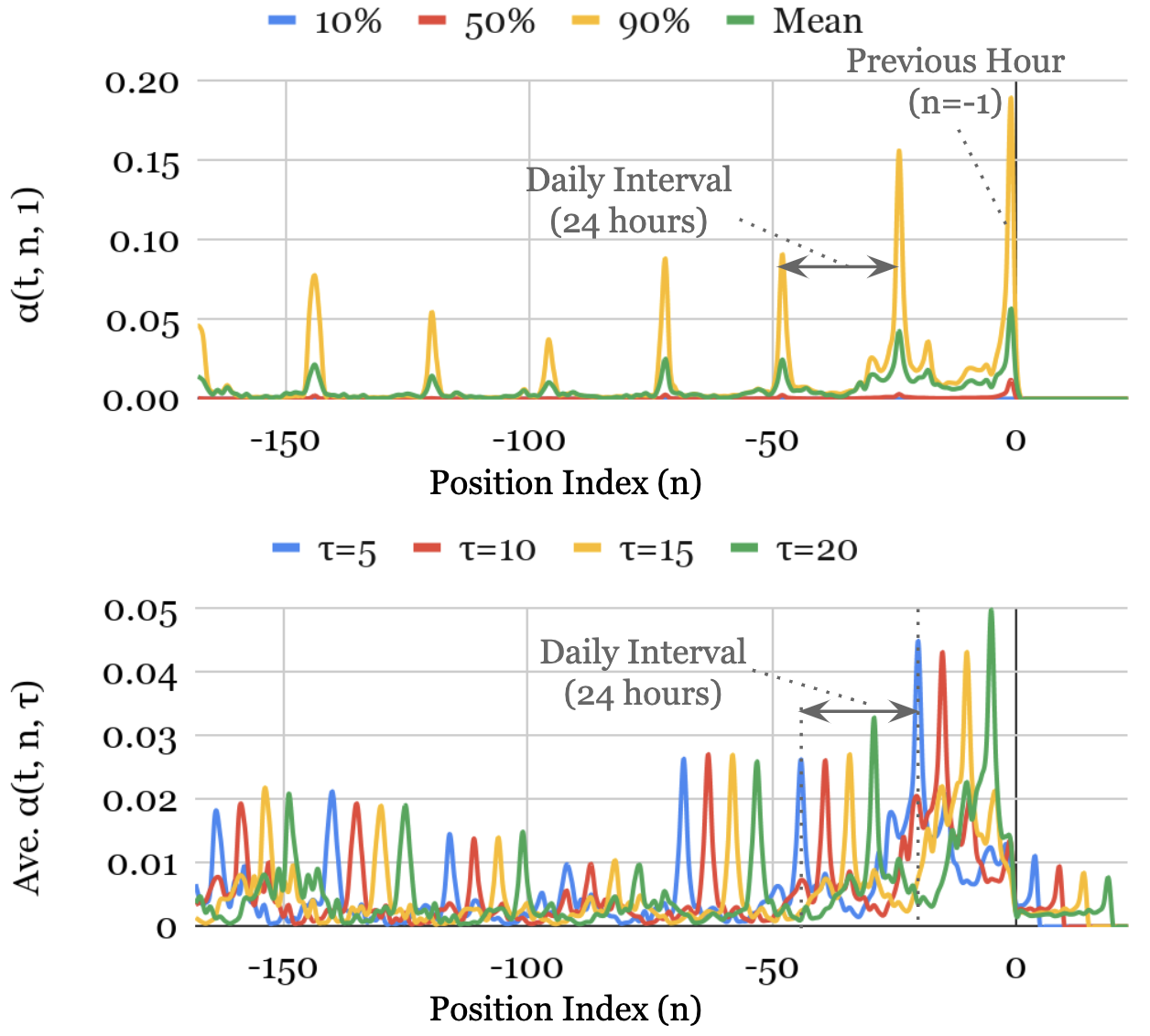}
    \caption{Electricity}
\end{subfigure} 
\begin{subfigure}{0.49\textwidth}
\centering
    \includegraphics[width=\linewidth]{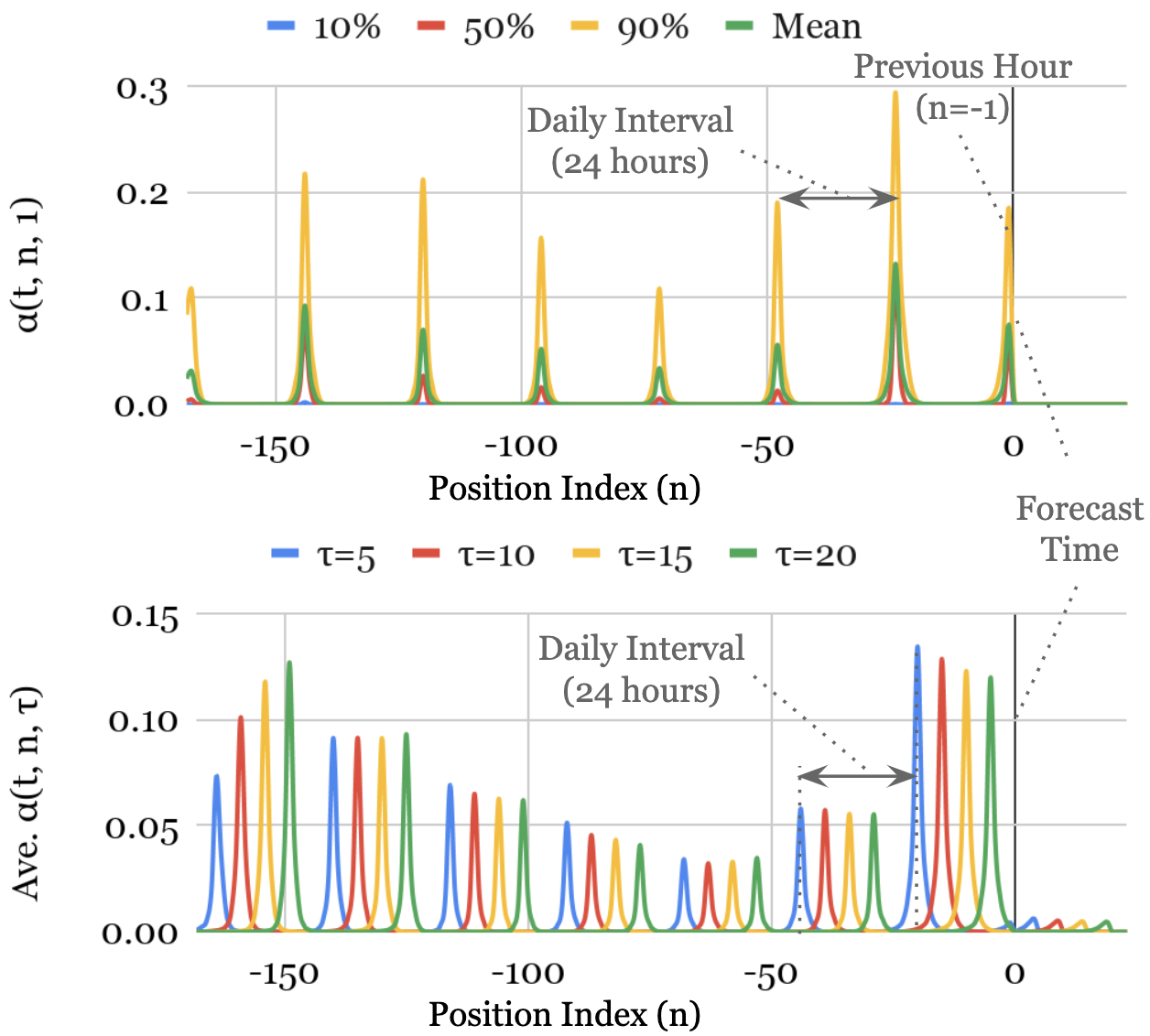}
        \caption{Traffic}
\end{subfigure} 
\vfill
\begin{subfigure}{0.49\textwidth}
\centering
    \includegraphics[width=\linewidth]{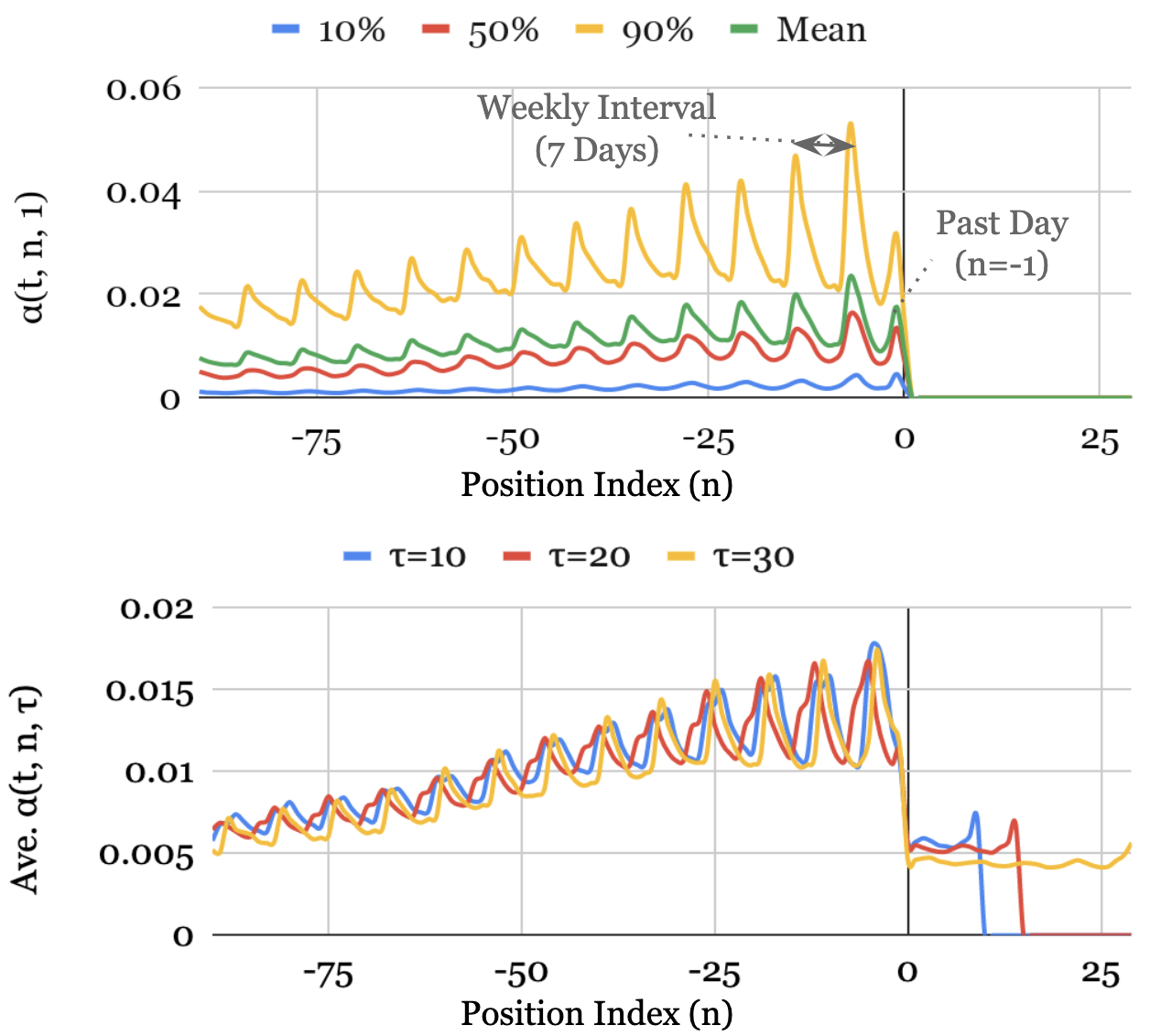}
    \caption{Retail}
\end{subfigure} 
\begin{subfigure}{0.49\textwidth}
\centering
\includegraphics[width=\linewidth]{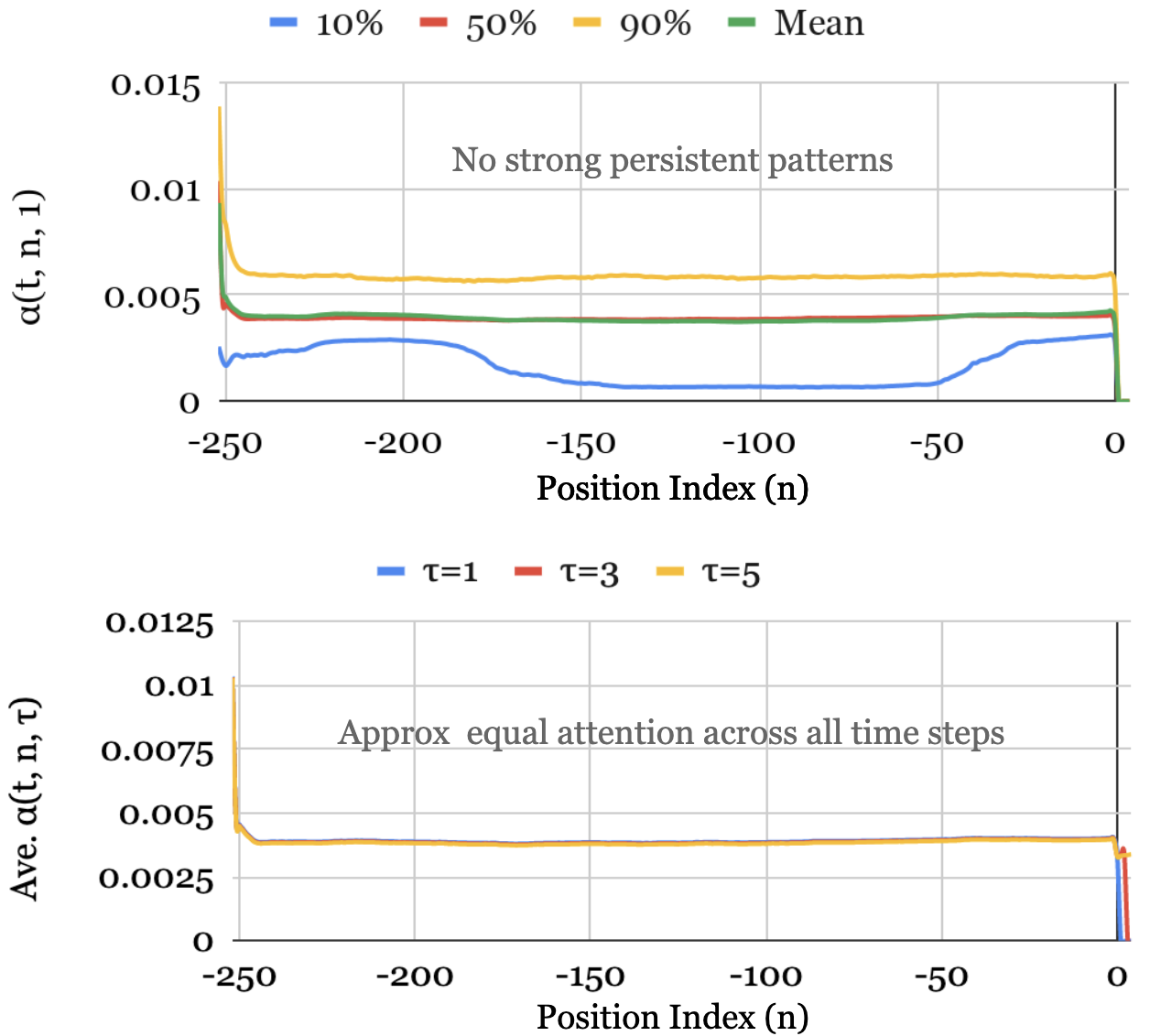}
\caption{Volatility}
\end{subfigure}
\caption{Persistent temporal patterns across datasets. Clear seasonality observed for the Electricity, Traffic and Retail datasets, but no strong persistent patterns seen in Volatility dataset. Upper plot -- percentiles of attention weights for one-step-ahead forecast. Lower plot -- average attention weights for forecast at various horizons.}
\label{fig:traffic_patterns}
\end{figure} 

Attention weight patterns can be used to shed light on the most important past time steps that the TFT model bases its decisions on. In contrast to other traditional and machine learning time series methods, which rely on model-based specifications for seasonality and lag analysis, the TFT can learn such patterns from raw training data.
%In traditional time-series machine learning, such past time step relevance determination is done via seasonality-trend decompositions, whereas TFT learns such patterns from raw training data.

Fig. \ref{fig:traffic_patterns} shows the attention weight patterns across all our test datasets -- with the upper graph plotting the mean along with the $10^{th}$, $50^{th}$ and $90^{th}$ percentiles of the attention weights for one-step-ahead forecasts (i.e. $\alpha(t, 1, \tau)$) over the test set, and the bottom graph plotting the average attention weights for various horizons (i.e. $\tau \in \{ 5, 10, 15, 20 \}$). We observe that the three datasets exhibit a seasonal pattern, with clear attention spikes at daily intervals observed for Electricity and Traffic, and a slightly weaker weekly patterns for Retail. For Retail, we also observe the decaying trend pattern, with the last few days dominating the importance.

No strong persistent patterns were observed for the Volatility -- attention weights equally distributed across all positions on average. This resembles a moving average filter at the feature level, and -- given the high degree of randomness associated with the volatility process -- could be useful in extracting the trend over the entire period by smoothing out high-frequency noise.
 
TFT learns these persistent temporal patterns from the raw training data without any human hard-coding. Such capability is expected to be very useful in building trust with human experts via sanity-checking. Model developers can also use these towards model improvements, e.g. via specific feature engineering or data collection.

\subsection{Identifying Regimes \& Significant Events}

Identifying sudden changes in temporal patterns can also be very useful, as temporary shifts can occur due to the presence of significant regimes or events. For instance, regime-switching behavior has been widely documented in financial markets \cite{RegimeSwitching}, with returns characteristics -- such as volatility -- being observed to change abruptly between regimes. 
As such, identifying such regime changes provides strong insights into the underlying problem which is useful for identification of the significant events.

Firstly, for a given entity, we define the average attention pattern per forecast horizon as:
\begin{equation}
\bar{\alpha}(n, \tau) = \sum\nolimits_{t=1}^{T} \alpha(t, j, \tau) / T,
\end{equation}
%
%\begin{equation}
%\bar{\alpha}(n, \tau) = \frac{\sum_{t=1}^{T} \alpha(t, n, \tau)}{\sum_{t=1}^{T} \sum_{j=-k}^{\tau_{max}} \alpha(t, j, \tau)} ,
%\end{equation}
%
and then construct $\bar{\bm{\alpha}}(\tau) = [\bar{\alpha}(-k, \tau), \dots, \bar{\alpha}(\tau_{max}, \tau)]^T$. To compare similarities between attention weight vectors, we use the distance metric proposed by \cite{BDistance}: 
\begin{align}
\kappa(\bm{p}, \bm{q}) = \sqrt{1-\rho(p,q)}, &
\end{align}
where $\rho(\bm{p},\bm{q}) = \sum_{j} \sqrt{p_j q_j}$ is the Bhattacharya coefficient \cite{BCoeff} measuring the overlap between discrete distributions -- with $p_j, q_j$ being elements of probability vectors  $\bm{p},\bm{q}$ respectively. For each entity, significant shifts in temporal dynamics are then measured using the distance between attention vectors at each point with the average pattern, aggregated for all horizons as below:
\begin{align}
\text{dist}(t) =  \sum\nolimits_{\tau=1}^{\tau_{max}} \kappa \big( \bar{\bm{\alpha}}(\tau), \bm{\alpha}(t, \tau) \big) / \tau_{max},
\end{align}
where $\bm{\alpha}(t, \tau) = [{\alpha}(t, -k, \tau), \dots, {\alpha}(t, \tau_{max}, \tau)]^T$. 

\begin{figure}[btp]
\includegraphics[width=1.0\linewidth]{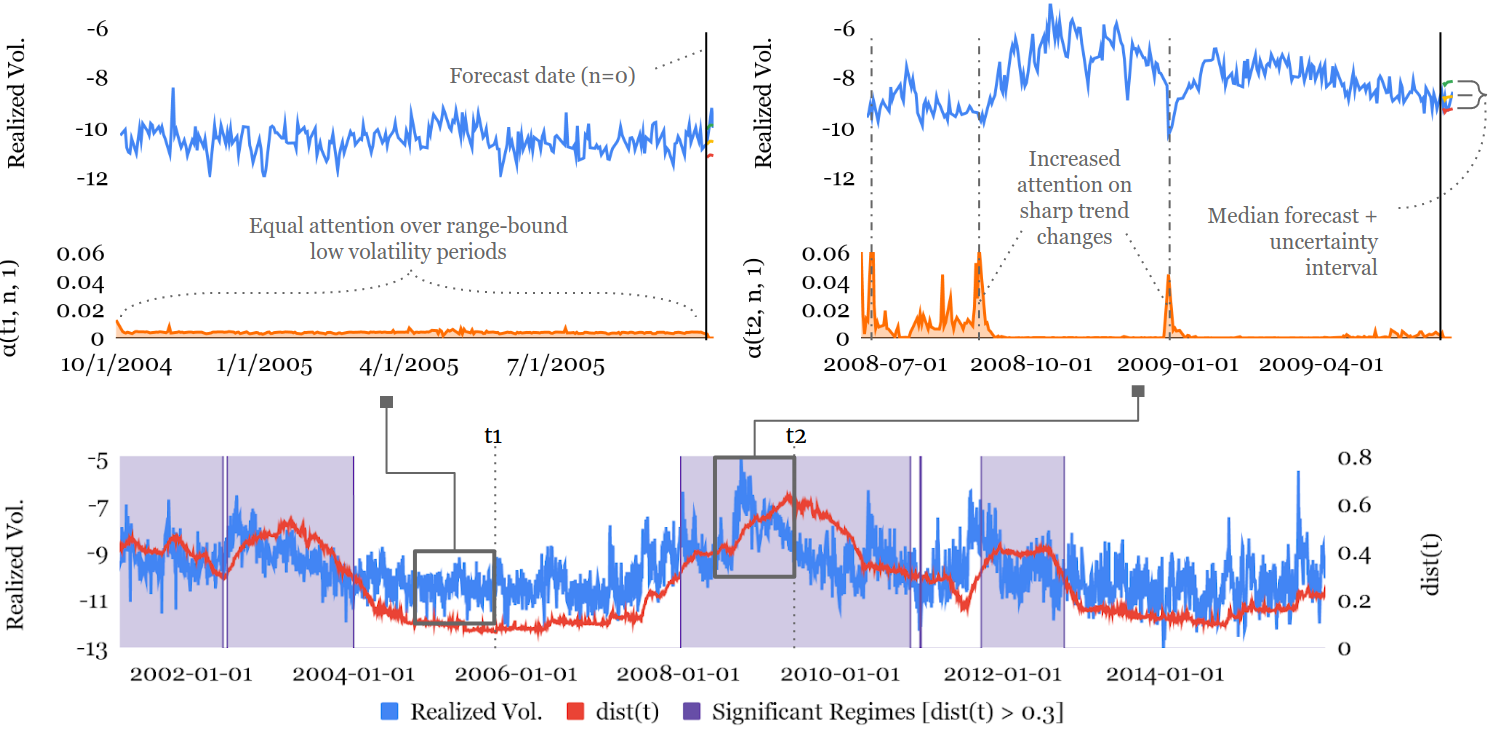}
\caption{Regime identification for S\&P 500 realized volatility. Significant deviations in attention patterns can be observed around periods of high volatility -- corresponding to the peaks observed in $\text{dist}(t)$. We use a threshold of $\text{dist}(t) > 0.3$ to denote significant regimes, as highlighted in purple. Focusing on periods around the 2008 financial crisis, the top right plot visualizes $\alpha(t, n, 1)$ midway through the significant regime, compared to the normal regime on the top left.}
\label{fig:vol_regime}
\end{figure}

Using the volatility dataset, we attempt to analyse regimes by applying our distance metric to the attention patterns for the S\&P 500 index over our training period (2001 to 2015). 
Plotting $\text{dist}(t)$ against the target (i.e. log realized volatility) in the bottom chart of Fig. \ref{fig:vol_regime}, significant deviations in attention patterns can be observed around periods of high volatility (e.g. the 2008 financial crisis) -- corresponding to the peaks observed in $\text{dist}(t)$.
From the plots, we can see that TFT appears to alter its behaviour between regimes -- placing equal attention across past inputs when volatility is low, while attending more to sharp trend changes during high volatility periods -- suggesting differences in temporal dynamics learned in each of these cases.

\section{Conclusions}
We introduce TFT, a novel attention-based deep learning model for interpretable high-performance multi-horizon forecasting. To handle static covariates, a priori known inputs, and observed inputs effectively across wide range of multi-horizon forecasting datasets, TFT uses specialized components. Specifically, these include: 
(1) sequence-to-sequence and attention based temporal processing components that capture time-varying relationships at different timescales, 
(2) static covariate encoders that allow the network to condition temporal forecasts on static metadata, 
(3) gating components that enable skipping over unnecessary parts of the network, 
(4) variable selection to pick relevant input features at each time step, and 
(5) quantile predictions to obtain output intervals across all prediction horizons. 
On a wide range of real-world tasks -- on both simple datasets that contain only known inputs and complex datasets which encompass the full range of possible inputs -- we show that TFT achieves state-of-the-art forecasting performance. 
Lastly, we investigate the general relationships learned by TFT through a series of interpretability use cases -- proposing novel methods to use TFT to 
(i) analyze important variables for a given prediction problem, 
(ii) visualize persistent temporal relationships learned  (e.g. seasonality), and
(iii) identify significant regimes changes.

\ifthenelse{\boolean{anon}}{~}{
\section{Acknowledgements}
The authors gratefully acknowledge discussions with Yaguang Li, Maggie Wang, Jeffrey Gu, Minho Jin and Andrew Moore that contributed to the development of this paper.
}
\newpage
\footnotesize{
\bibliography{fusion}}

\newpage
\section*{APPENDIX}
\setcounter{section}{1}
\appendix

\section{Dataset and Training Details}
\label{apdx:dataset_info}
We provide all the sufficient information on feature pre-processing and train/test splits to ensure reproducibility of our results.

\textbf{Electricity:} Per \cite{DeepAR}, we use 500k samples taken between 2014-01-01 to 2014-09-01 -- using the first $90\%$ for training, and the last $10\%$ as a validation set. Testing is done over the 7 days immediately following the training set -- as described in \cite{DeepAR, MatrixFactorisation}. Given the large differences in magnitude between trajectories, we also apply z-score normalization separately to each entity for real-valued inputs. In line with previous work, we consider the electricity usage, day-of-week, hour-of-day and and a time index -- i.e. the number of time steps from the first observation -- as real-valued inputs, and treat the entity identifier as a categorical variable.

\textbf{Traffic:}  Tests on the Traffic dataset are also kept consistent with previous work, using 500k training samples taken before 2008-06-15 as per \cite{DeepAR}, and split in the same way as the Electricity dataset. For testing, we use the 7 days immediately following the training set, and z-score normalization was applied across all entities. For inputs, we also take traffic occupancy, day-of-week, hour-of-day and and a time index as real-valued inputs, and the entity identifier as a categorical variable.

\textbf{Retail:} We treat each product number-store number pair as a separate entity, with over 135k entities in total. The training set is made up of 450k samples taken between 2015-01-01 to 2015-12-01, validation set of 50k samples from the 30 days after the training set, and test set of all entities over the 30-day horizon following the validation set. We use all inputs from the Kaggle competition.
%(please see Appendix \ref{apdx:interp_variable} for full list -- including additional variables for the day-of-week, day-of-month, and month
Data is resampled at regular daily intervals, imputing any missing days using the last available observation. We include an additional 'open' flag to denote whether data is present on a given day. We group national, regional, and local holidays into separate variables. We apply a log-transform on the sales data, and adopt z-score normalization across all entities. We consider log sales, transactions, oil to be real-valued and the rest to be categorical.

\textbf{Volatility:} We use the data from 2000-01-03 to 2019-06-28 -- with the training set consisting of data before 2016, the validation set from 2016-2017, and the test set data from 2018 onwards. For the target, we focus on 5-min sub-sampled realized volatility (i.e. the rv5\_ss column ), and add daily open-to-close returns as an extra exogenous input. Additional variables are included for the day-of-week, day-of-month, week-of-year, and month -- along with a 'region' variable for each index (i.e. Americas, Europe or Asia). Finally, a time index is added to denote the number of days from the first day in the training set. We treat all date-related variables (i.e. day-of-week, day-of-month, week-of-year, and month) and the region as categorical inputs. A log transformation is applied to the target, and all inputs are z-score normalized across all entities.

\section{Interpretability Results}
\label{apdx:interp}

Apart from Sec. \ref{sec:interp}, which highlights our most prominent findings, we present the remaining results here for completeness.
 
\subsection{Variable Importance}
\label{apdx:interp_variable}
Table \ref{tab:apdx_varselect} shows the variable importance scores for the remaining Electricity, Traffic and Volatility datasets. As these datasets only have one static input, the network allocates full weight to the entity identifier for Electricity and Traffic, along with the region input for Volatility. We also observe two types of important time-dependent inputs -- those related to past values of the target as before, and those related to calendar effects. For instance, the hour-of-day plays a significant roles for Electricity and Traffic datasets, echoing the daily seasonality observed in the next section. In the Volatility dataset, the day-of-month is observed to play a significant role in future inputs -- potentially reflecting turn-of-month effects \cite{TurnOfMonth}.
\vfill
\begin{table}[bh]
\caption{Variable importance scores for the Electricity, Traffic and Volatility datasets. The most significant variable of each input category is highlighted in purple. As before, past values of the target play a significant role -- being the top 1 or 2 most significant past input across datasets. The role of seasonality can also be seen in Electricity and Traffic, where the past and future values of the hour-of-day is important for forecasts.}
\label{tab:apdx_varselect}
\centerline{
\begin{subtable}[t]{0.42\textwidth}
\hspace{-0.5cm}
\begin{tabular}{@{}llll@{}}
\toprule
\textbf{}                    & \textbf{10\%} & \textbf{50\%} & \textbf{90\%} \\ \midrule
{\ul \textbf{Static}} &               &               &               \\
ID                           & \textcolor{violet}{\textbf{1.000}}             & \textcolor{violet}{\textbf{1.000}}             & \textcolor{violet}{\textbf{1.000}}             \\ \midrule
{\ul \textbf{Past}}   &               &               &               \\
Hour of Day                  & \textcolor{violet}{\textbf{0.437}}         & \textcolor{violet}{\textbf{0.462}}         & \textcolor{violet}{\textbf{0.473}}         \\
Day of Week                  & 0.078         & 0.099         & 0.151         \\
Time Index                   & 0.066         & 0.077         & 0.092         \\
Power Usage                  & 0.342         & 0.359         & 0.366         \\  \midrule
{\ul \textbf{Future}} &               &               &               \\
Hour of Day                  & \textcolor{violet}{\textbf{0.718}}         & \textcolor{violet}{\textbf{0.738}}         & \textcolor{violet}{\textbf{0.739}}         \\
Day of Week                  & 0.109         & 0.124         & 0.166         \\
Time Index                   & 0.114         & 0.137         & 0.155         \\  \bottomrule
\end{tabular}
\caption{Electricity}
\end{subtable} \hfill
\begin{subtable}[t]{0.42\textwidth}
\hspace{-0.5cm}
\begin{tabular}{@{}llll@{}}
\toprule
\textbf{}                    & \textbf{10\%} & \textbf{50\%} & \textbf{90\%} \\ \midrule
{\ul \textbf{Static}} &               &               &               \\
ID                           & \textcolor{violet}{\textbf{1.000}}             & \textcolor{violet}{\textbf{1.000}}             & \textcolor{violet}{\textbf{1.000}}             \\ \midrule
{\ul \textbf{Past}}   &               &               &               \\
Hour of Day                  & 0.285         & 0.296         & 0.300         \\
Day of Week                  & 0.117         & 0.122         & 0.124         \\
Time Index                   & 0.107         & 0.109         & 0.111         \\
Occupancy                    & \textcolor{violet}{\textbf{0.471}}         & \textcolor{violet}{\textbf{0.473}}         & \textcolor{violet}{\textbf{0.483}}         \\  \midrule
{\ul \textbf{Future}} &               &               &               \\
Hour of Day                  & \textcolor{violet}{\textbf{0.781}}         & \textcolor{violet}{\textbf{0.781}}         & \textcolor{violet}{\textbf{0.781}}         \\
Day of Week                  & 0.099         & 0.100         & 0.102         \\
Time Index                   & 0.117         & 0.119         & 0.121         \\   \bottomrule
\end{tabular}
\caption{Traffic}
\end{subtable}}
\end{table}
\begin{table}[p]
\centering
\ContinuedFloat
\begin{subtable}[t]{0.5\textwidth}
\begin{tabular}{@{}llll@{}}
\toprule
\textbf{}                    & \textbf{10\%} & \textbf{50\%} & \textbf{90\%} \\ \midrule
{\ul \textbf{Static}} &               &               &               \\
Region                       & \textcolor{violet}{\textbf{1.000}}             & \textcolor{violet}{\textbf{1.000}}             & \textcolor{violet}{\textbf{1.000}}             \\ \midrule
{\ul \textbf{Past}}   &               &               &               \\
Time Index                   & 0.093         & 0.098         & 0.142         \\
Day of Week                  & 0.003         & 0.004         & 0.004         \\
Day of Month                 & 0.017         & 0.027         & 0.028         \\
Week of Year                 & 0.022         & 0.057         & 0.068         \\
Month                        & 0.008         & 0.009         & 0.011         \\
Open-to-close Returns        & 0.078         & 0.158         & 0.178         \\
Realised Vol                 & \textcolor{violet}{\textbf{0.620}}         & \textcolor{violet}{\textbf{0.647}}         & \textcolor{violet}{\textbf{0.714}}        \\ \midrule
{\ul \textbf{Future}} &               &               &               \\
Time Index                   & 0.011         & 0.014         & 0.024         \\
Day of Week                  & 0.019         & 0.072         & 0.299         \\
Day of Month                 & \textcolor{violet}{\textbf{0.069}}         & \textcolor{violet}{\textbf{0.635}}         & \textcolor{violet}{\textbf{0.913}}         \\
Week of Year                 & 0.026         & 0.060         & 0.227         \\
Month                        & 0.008         & 0.055         & 0.713         \\ \bottomrule
\end{tabular}
\caption{Volatility}
\end{subtable}
\end{table}

\end{document}